\documentclass[]{IEEEtran}
\usepackage{amsmath,amsfonts}
\usepackage{algorithmic}
\usepackage{algorithm}
\usepackage{array}
\usepackage{textcomp}
\usepackage{stfloats}
\usepackage{url}
\usepackage{verbatim}
\usepackage{graphicx}
\usepackage{cite}
\usepackage{xspace}
\usepackage{bm}

\usepackage[colorlinks,linkcolor = black,anchorcolor = black,citecolor = black]{hyperref}
\usepackage{epstopdf}
\usepackage[labelformat=simple]{subcaption}

\usepackage{diagbox}
\usepackage{tabularx}
\usepackage{subcaption}
\usepackage{amssymb}

\usepackage{color,xcolor}

\newcommand{\name}{Meta-SimGNN\xspace}

\begin{document}

\title{Meta-SimGNN: Adaptive and Robust WiFi Localization Across Dynamic Configurations and Diverse Scenarios}

\author{Qiqi Xiao, Ziqi Ye, Yinghui He, \IEEEmembership{Member,~IEEE}, Jianwei Liu, and Guanding Yu, \IEEEmembership{Senior Member,~IEEE}

\thanks{Manuscript received 12 May 2025; revised 23 September 2025; accepted 16 November 2025.
{\it(Corresponding author: Yinghui He)}
}

\thanks{Q. Xiao, Z. Ye, Y. He, and G. Yu are with the College of Information Science and Electronic Engineering, Zhejiang University, Hangzhou 310027, China (email: \{xiaoqiqi, yeziqi, 2014hyh, yuguanding\}@zju.edu.cn).

J. Liu is with the College of Information Science and Electronic Engineering, Zhejiang University, Hangzhou 310027, China, and also with the School of Information and Electrical Engineering, Hangzhou City University, Hangzhou 310015, China (email: jianweiliu@zju.edu.cn).}}


\maketitle

\begin{abstract}
To promote the practicality of deep learning-based localization, existing studies aim to address the issue of scenario dependence through meta-learning. However, these studies primarily focus on variations in environmental layouts while overlooking the impact of changes in device configurations, such as bandwidth, the number of access points (APs), and the number of antennas used.
Unlike environmental changes, variations in device configurations affect the dimensionality of channel state information (CSI), thereby compromising  neural network usability.
To address this issue, we propose \name, a novel WiFi localization system that integrates graph neural networks with meta-learning to improve localization generalization and robustness.
First, we introduce a fine-grained CSI graph construction scheme, where each AP is treated as a graph node, allowing for adaptability to changes in the number of APs.
To structure the features of each node, we propose an amplitude-phase fusion method and a feature extraction method. The former utilizes both amplitude and phase to construct CSI images, enhancing data reliability, while the latter extracts dimension-consistent features to address variations in bandwidth and the number of antennas.
Second, a similarity-guided meta-learning strategy is developed to enhance adaptability in diverse scenarios.
The initial model parameters for the fine-tuning stage are determined by comparing the similarity between the new scenario and historical scenarios, facilitating rapid adaptation of the model to the new localization scenario.
Extensive experimental results over commodity WiFi devices in different scenarios show that \name outperforms the baseline methods in terms of localization generalization and accuracy.
\end{abstract}


\section{Introduction}\label{Introduction}
\IEEEPARstart{W}{ith} the rapid development of Internet of Things technology, various intelligent applications, such as augmented reality, industrial mobile robots, and autonomous driving, are emerging~\cite{chettri2019comprehensive,sinche2019survey}. These applications heavily rely on indoor localization technologies, drawing significant attention from both industry and academia~\cite{li2020toward,shit2018location}.
However, the global positioning system, the most commonly used localization solution, is difficult to achieve precise localization indoors due to signal obstruction caused by buildings~\cite{lin2016enhanced,gu2019indoor}.
To address this issue, other wireless signals, such as Bluetooth, WiFi, and ultra-wideband, have been explored for indoor localization~\cite{zafari2019survey,gu2019indoor,he2024forward}.
Among these, WiFi stands out as the most promising option, thanks to its widespread deployment and immediate availability~\cite{laoudias2018survey} in indoor scenarios.

Mainstream WiFi localization solutions can be divided into two types:  modeling-based and learning-based approaches~\cite{zhu2020indoor}. Modeling-based approaches primarily achieve localization by estimating ranging and angular measurements, such as time of flight, angle of arrival (AoA), and angle of departure (AoD), with channel state information (CSI) at the WiFi receiver and further determining the location with geometric relationships.
However, they heavily depend on the line-of-sight (LoS) path and thus generally perform poorly in the non-line-of-sight (NLoS) scenarios~\cite{zhang2019efficient,stahlke2023indoor}.
In contrast, learning-based approaches leverage powerful deep learning (DL) methods and are not limited to LoS scenarios~\cite{wang2022framework,hu2024cross}. Thus, they hold greater practical value in real-world environments characterized by rich multipath effects and the potential absence of LoS path, compared to modeling-based approaches~\cite{huang2020machine}.

However, the success of DL-based methods heavily relies on extensive high-quality CSI~\cite{he2023ai}, and WiFi CSI is determined by the environment as it contains the path information of signal transmission.
When a model trained on existing scenarios is applied to a new and unseen scenario, it often performs poorly due to the lack of exposure to data from the new scenario.
A straightforward solution is to re-collect data and re-train the model, causing high overheads.
To address it, meta-learning has been adopted since it develops a generalized initialization that facilitates rapid adaptation to unseen scenarios by training the model on a variety of scenarios~\cite{wei2023meta,owfi2023meta,jiao2024dynamic,pu2023bayesian}.
Nevertheless, prior works typically focus on deployment variation across different environments, e.g., the locations of the access points (APs) and surrounding reflectors~\cite{zhang2019efficient,li2019af,zhu2022intelligent,stahlke2023indoor}, while overlooking the impact of changes in device configurations.
In fact, the configurations of the transceiver devices (e.g., the number, bandwidth, and activated antennas of the WiFi APs) 
generally are different across different scenarios, and even dynamically vary in the same scenario for communication purposes.
Different from environmental changes, the variation in the configuration would result in different WiFi CSI dimensions and contained information.
Such variation is challenging to address with existing solutions, as it necessitates adjustments to the model structure, which conflicts with the fixed input dimensions required by these methods. This limitation significantly hampers the practical deployment of WiFi localization systems.

To address this issue, we propose a novel WiFi localization system, namely \name, in this paper.
It leverages the scalability and topology-invariance properties of graph neural networks (GNNs)~\cite{wu2020comprehensive} to effectively handle variations in the number of WiFi APs, while incorporating the fast adaptation of meta-learning to achieve efficient cross-scenario localization.
By collecting a few CSI samples in different scenarios and using the CSI corresponding to each AP as the input of the graph node to train and fine-tune the GNN, \name can achieve robust cross-scenario localization at low cost.

However, we still face two challenges.
First, the bandwidth of each AP and the number of activated transmit/receive antennas may dynamically fluctuate due to interference from surrounding devices, and thus, the size of the CSI for each AP may still change, making it difficult to be directly input into the same GNN. 
In addition, the amplitude and phase of the CSI contain different features, with their noise and errors differing in both origin and degree. Directly inputting the raw CSI into the GNN may result in poor localization performance.
Second, existing meta-learning strategies are primarily designed for cross-task problems and have been extensively validated on structured data such as images.
However, localization is not a cross-task problem since the task remains the same, but rather a cross-scenario problem, where variations arise from the differences in data distributions across environments.
Existing works of using meta-learning for localization fail to account for these specific characteristics, which may lead to poor adaptation.

To address the first challenge, we propose a fine-grained CSI graph construction scheme consisting of an amplitude-phase fusion method and a feature extraction method. The former first denoises and eliminates errors in the amplitude and phase separately. Then, it constructs a CSI image by combining them in a 2:1 ratio under the different reliability of amplitude and phase, thereby improving the localization robustness. The latter extracts bandwidth-independent and dimension-consistent features with spatial pyramid pooling~\cite{he2015spatial} and uses them as graph node inputs for the GNN, thereby resolving the dimensionality issue. 
To address the second issue, we observe that localization environments often contain several typical scenarios, and most new environments share a certain degree of similarity with one of them. Motivated by this observation, we design a similarity-guided meta-learning strategy to enhance adaptability.
During the meta-training stage, the model is trained on multiple representative scenarios.
For a new scenario, the initial model parameters are selected based on the similarity between the new and historical scenarios. This allows the fine-tuned model to quickly and accurately estimate locations in the new scenario.
We also build a prototype of \name and conduct real-world experiments in three scenarios to evaluate its performance. The results indicate that \name has superior localization generalization and robustness compared to the baseline methods.
The main contributions can be summarized as follows.

$\bullet$  We propose \name, a practical WiFi localization solution that rapidly adapts to dynamic device configurations and diverse scenarios while enhancing localization robustness.

$\bullet$  We propose a fine-grained CSI graph construction scheme. Specifically, an amplitude-phase fusion method is designed for constructing CSI images and enhancing data reliability, and a feature extraction method is designed for extracting bandwidth-independent and dimension-consistent features.

$\bullet$ We innovate a similarity-guided meta-learning strategy to enhance adaptability. It determines the initial model parameters for the fine-tuning stage using the similarity between the new scenario and the existing scenarios.

$\bullet$ We implement \name with commodity WiFi devices and evaluate its performance across different scenarios. The experiment results show that \name rapidly adapts to real scenarios and shows superior performance compared to the baseline methods.

The rest of this paper is organized as follows. Section~\ref{Related Work} reviews the related works. The system model and preliminaries are introduced in Section~\ref{System Model}. Section~\ref{Overview of Meta-SimGNN}
presents an overview of the proposed \name.
The fine-grained GNN construction and the similarity-guided meta-learning strategy are described in detail in Sections~\ref{CSI Graph Construction} and~\ref{ML-GDS}, respectively. Section~\ref{Experimental Setup} explains \name's implementation and
Section~\ref{Experiment Evaluation} presents the performance of \name.
Finally, the whole paper is concluded in Section~\ref{Conclusion}.

\section{Related Work}\label{Related Work}
Indoor localization techniques have been studied for many years, including both modeling-based and learning-based approaches.
Existing modeling-based methods~\cite{kotaru2015spotfi, chintalapudi2010indoor} typically utilize received signal strength and/or CSI to estimate distance and angle for localization, achieving high localization accuracy under favorable conditions (e.g., high signal-to-noise ratio, multiple antennas, and LoS). However, they generally perform poorly in the NLoS scenarios.
In comparison, learning-based approaches are generally more attractive~\cite{zhu2020indoor,nessa2020survey}.
Many existing works~\cite{chen2017confi,wang2018deep,foliadis2021csi,zhang2024tomfi,ayyalasomayajula2020deep} have realized a higher performance by employing convolutional neural networks (CNNs).
For example, the authors in~\cite{ayyalasomayajula2020deep} have proposed DLoc, which learns implicit representation relationships between wireless signals and ground truth locations, achieving high accuracy. 
Nevertheless, its superior performance depends on large training datasets. Although the MapFind platform has been developed to automate data collection, it is still difficult to cover all possible environmental conditions.
Considering that the environment changes dynamically (e.g., caused by people walking around), CSI easily becomes outdated, and existing studies provide several solutions to address it.
Data augmentation, one of the representative technologies, aims to enhance training datasets by artificially generating new data samples.
For example, a cellular localization method incorporates augmented data to improve localization accuracy~\cite{rizk2019effectiveness}.
Another promising solution is incremental learning, such as the intelligent localization scheme in~\cite{zhu2022intelligent} without retraining, greatly reducing the offline training time.

Most of the above studies focus on single-AP scenarios. However, single-AP localization may fail when obstacles block the signal path between the AP and the user equipment (UE). 
To address this limitation, multi-AP localization systems have been widely investigated, as they provide richer spatial information and significantly improve system robustness. 
Nonetheless, efficiently integrating such heterogeneous data remains a challenge.
In recent years, GNN has been introduced to multi-AP localization owing to its inherent advantage in processing graph-structured data, effectively addressing the challenge of varying AP numbers~\cite{lezama2021indoor, yan2021graph,wang2024graph,zhang2025gloc,yan2025attentional}.
GNN constructs the graph of APs and their relationships, and employs message passing and aggregation mechanisms that allow each AP to integrate information from its neighbors, thereby fully exploiting the topological relationships between APs to improve localization performance. 
The authors in~\cite{wang2024graph} have proposed a localization model that combines the AP selection network and the graph-based location mapping network to improve both localization accuracy and efficiency. 
In~\cite{yan2025attentional}, the authors have proposed an attention-based GNN model to adaptively learn an optimal adjacency matrix, thereby reducing computational complexity in large-scale localization and improving robustness in NLoS conditions.
However, these methods largely overlook the impact of variations in AP configurations (e.g., bandwidth and number of antennas).

In addition, most current studies focus on training and testing within the same scenario, resulting in significant performance degradation when the trained models are applied to new scenarios.
Therefore, it is expected to design a universal localization method that can generalize across scenarios, as required by 3GPP TR 38.843 standard~\cite{3rd2023}.
Two main approaches have been explored: transfer learning (TL) and meta-learning. 
The TL technique aims to leverage knowledge from a source task to improve the learning performance on a target task.
For example, CRISLoc~\cite{gao2020crisloc} has exploited TL to reconstruct a CSI database by combining outdated fingerprints with a few new measurements for localization.
However, TL-based methods may require extensive preprocessing to maintain good performance~\cite{liu2017toward}, particularly when source and target domains exhibit significant differences in large indoor scenarios. This requires tremendous new samples to update the whole target region, otherwise negative transfer may occur, limiting their scalability.
Meta-learning, by contrast, adopts a more flexible paradigm that focuses on learning itself, i.e., learning to learn. It enables models to quickly adapt to new scenarios after fine-tuning using some newly collected data~\cite{finn2017model}.
Recent meta-learning-based studies~\cite{wei2023meta,owfi2023meta,jiao2024dynamic,pu2023bayesian} generally employ task-agnostic meta-learning frameworks directly, without fully considering the unique nature of localization tasks.
The authors in~\cite{chen2023fast} have proposed a federated meta-learning framework to rapidly adapt to new environments, but the communication cost is high.
MetaLoc~\cite{gao2023metaloc}, employing the model-agnostic meta-learning (MAML) for localization, utilized both centralized and distributed paradigms for fast adaptation. 
Nevertheless, the framework based on MAML and CNN does not consider the impact of changes in input dimensions arising from variations in device configurations.

\section{Preliminary and Motivation}\label{System Model}
In this section, we first introduce the foundations of WiFi localization, i.e., the CSI model. Then, we introduce the preliminaries of meta-learning and GNN, and explain the motivation of adopting them to implement \name.

\subsection{CSI Model}\label{CSI Model}
Existing WiFi localization solutions mainly rely on the CSI measured from received pilots contained in WiFi packets. Specifically, the pilots transmitted by the AP arrive at the UE after undergoing several different paths, and the UE measures the CSI by comparing the transmit pilots and the received signals. Without loss of generality, we consider $N$ WiFi APs around the UE with  $N^{\mathrm{tx}}_n$ and  $N^{\mathrm{rx}}$ being the number of antennas for the $n$-th WiFi AP and the UE, respectively.
The bandwidth occupied by the $n$-th WiFi AP is $BW_n$ and is divided into $K_n$ subcarriers since the orthogonal frequency division multiplexing technique is adopted.
Assuming $L$ paths, the CSI between the $n$-th WiFi AP and the UE at the $k$-th subcarrier can be represented as
\begin{equation} \label{eq:csi}
    \bm{H}_{n,k} = \sum_{l=1}^{L} \alpha_le^{-jd_l/\lambda_k} \bm{a}(\theta_l) \bm{a}(\phi_l)^H \in \mathbb{C}^{N^{\mathrm{rx}}\times N^{\mathrm{tx}}_n},
    \vspace{-1ex}
\end{equation}
where $\alpha_l$ is the path attenuation, $d_l$ is the path distance, $\theta_l$ is the AoD, and $\phi_l$ is the AoA with $\bm{a}(\cdot)$ being the steering vector. Moreover, $\lambda_k$ is the wavelength of the $k$-th subcarrier.

From~\eqref{eq:csi}, it can be seen that CSI contains information on the location of the UE, and the amplitude and phase can be extracted for localization.
Traditional localization methods (e.g., MUSIC algorithm~\cite{schmidt1986multiple,gupta2015music,zheng2018exploiting}) aim to extract the AoA and distance of the LoS path for estimating the UE's location. However, they fail to work in the case without the LoS path.
To address this issue, we aim to adopt the DL-based method.
The proposed \name employs DL techniques to fully exploit the localization features in CSI for accurate estimation.

\subsection{Meta-Learning Technique}\label{Meta-Learning Technique}
Although DL-based methods hold the potential for achieving higher performance, they also face challenges in adaptability~\cite{wang2015deepfi}.
The deep neural networks, trained using the CSI collected in one scenario, generally do not perform well in the other scenarios. Re-collecting a large amount of CSI and re-training the model to maintain good localization performance would cause a high maintenance cost. To tackle this problem, existing works~\cite{gao2023metaloc,jiao2024dynamic,wei2023meta} adopt the meta-learning technique that only needs fine-tuning on trained neural networks using some CSI samples collected from the new scenario.

Specifically, meta-learning mainly consists of two stages: meta-training and meta-testing (i.e., fine-tuning).
In the meta-training stage, the objective is to learn a set of well-generalized meta-parameters from multiple existing scenarios, which serves as an optimal initialization for fast adaptation in the meta-testing stage.
This can be achieved by adopting a double-loop update framework consisting of an inner loop and an outer loop. The inner loop is designed to enable rapid task-specific adaptation, while the outer loop aims to update the shared meta-parameters for better generalization across tasks.
Considering the CSI datasets measured from $P$ existing scenarios, the dataset of each scenario $\mathcal{T}_p$ is divided into a support set $\mathcal{S}_p$ for inner-loop training and a query set $\mathcal{Q}_p$ for outer-loop validation, i.e., $\mathcal{T}_p=[\mathcal{S}_p, \mathcal{Q}_p]$.
The model is trained to minimize the aggregate loss over the corresponding query sets $\mathcal{Q}_p$ after gradient updates on $\mathcal{S}_p$, thus improving the localization performance.
This process encourages the meta-parameters to capture transferable knowledge across all training problems, providing an initialization that is rapidly adaptable to the new problem. In the original meta-learning design, the problem corresponds to the task, while in our work, it specifically refers to the scenario.
During the meta-testing stage for the new scenario, these pre-trained meta-parameters serve as the initial model weights. The model is then fine-tuned with a few CSI samples measured from the new scenario, requiring only a few gradient descent steps. 
Consequently, the model efficiently specializes its general knowledge to the new scenario, thereby enabling fast and accurate localization.

However, existing meta-learning-based localization methods only focus on environmental changes. In fact, the heterogeneity of WiFi APs in terms of number and bandwidth also brings the different CSI and its size even in the same environment.
This cannot be directly solved using meta-learning since most consistent neural networks require consistent input in terms of size. To address this, we aim to explore the novel GNN as it can adapt to different input sizes.

\subsection{Graph Neural Network}\label{Graph Neural Network}
\begin{figure}[t]
  \centering
  \includegraphics[width=0.48\textwidth]{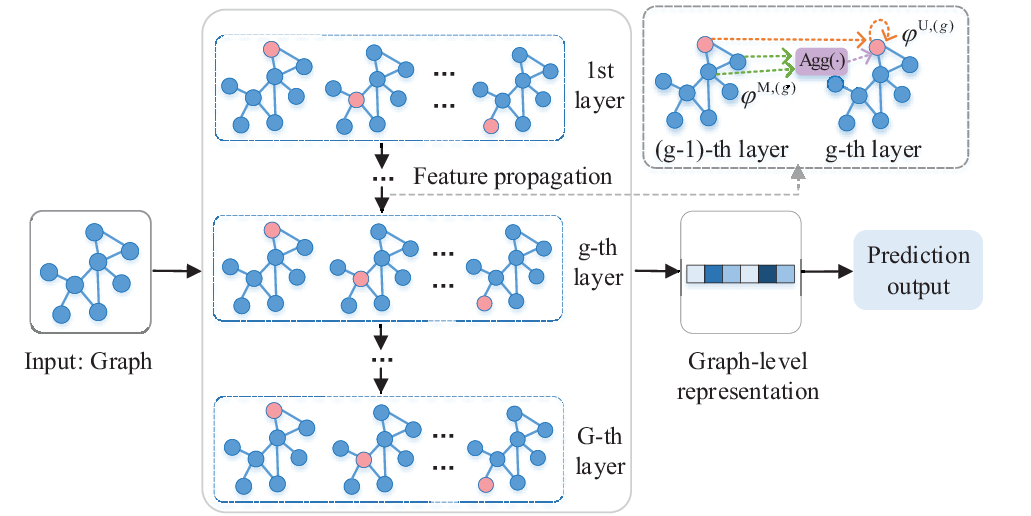}\\
  \caption{The structure of GNN.}
  \label{GNN_network}
  \vspace{-2ex}  
\end{figure}

\begin{figure*}[htbp]
  \centering
  \includegraphics[width=0.90\textwidth]{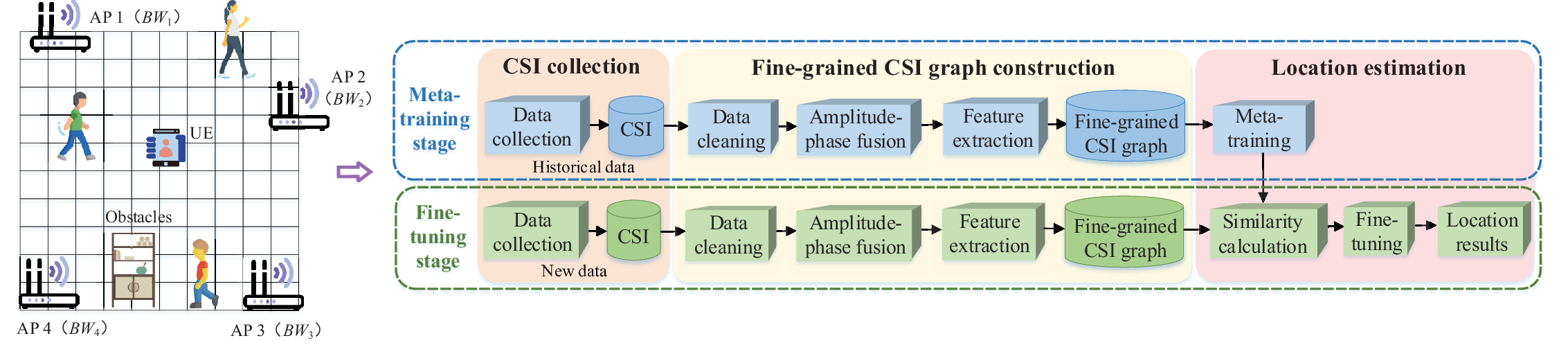}
  \vspace{-2ex}
  \caption{The framework of \name.}
  \label{Meta-SimGNN}
  \vspace{-3ex}
\end{figure*}

As shown in Fig.~\ref{GNN_network}, the GNN is a machine learning model that processes graph-structured data, which achieves great performance~\cite{wu2020comprehensive,buffelli2020meta}. It mainly consists of three modules: graph construction, feature propagation, and prediction output, detailed in the following.

\textbf{Graph construction.} Generally, a graph can be denoted as $\mathcal{G}=(\mathcal{V},\mathcal{E},\mathcal{X})$, where $\mathcal{V}$ is the set of nodes, $\mathcal{E}$ is the set of edges, and $\mathcal{X}$ is the node feature of graph~\cite{wu2020comprehensive}.
Assume there are $N$ nodes, i.e., $\mathcal{V}=\{v_{1}, v_{2}, ...,v_{N}\}$, and $e_{n,m}=(v_{n},v_{m})\in \mathcal{E}$ is the edge from $v_{n}$ to $v_{m}$.
The neighborhood of the node $v$ can be defined as $\mathcal{N}(v)=\{u\in \mathcal{V} \vert (v,u) \in \mathcal{E}\}$. Initially, the node feature of $v_{n}$ can be denoted as $\bm{x}^{(0)}_{n} \in \mathcal{X}$. The connection of the graph can be represented by an adjacency matrix $\bm{A} \in \mathbb{R}^{N \times N}$ as
\begin{equation}\label{eq2}
  \bm{A}_{n,m}=
  \left \{
  \begin{aligned}
    &w_{n,m},~{\rm{if}}~n\in\mathcal{N}(v), \\
    &0,~\rm otherwise,
  \end{aligned}
  \right.
\end{equation}
where $w_{n,m}$ is the edge feature (i.e., weight) of edge $e_{n,m}$.
Moreover, there is a special graph, i.e., an undirected graph, where the edges have no specified direction, and the adjacency matrix is symmetrical.

\textbf{Feature propagation.} After completing the graph construction, we need to learn the feature representation using graph convolutional layers. Let $G$ denote the number of graph convolutional layers, and the feature of the $n$-th input node at the $g$-th layer is $\bm{s}^{(g)}_{n}$ with the initial feature being $\bm{x}^{(0)}_{n}$. The detailed steps of feature propagation include message passing, aggregation, and feature update.
\begin{enumerate}
\item[$\bullet$] Message passing. In the $g$-th layer, the feature of $v_{n}$ and all its adjacent nodes in the $(g-1)$-th layer and the edge feature $w_{n,m}$ from $v_{n}$ to $v_{m}$ are used to generate messages to update the feature of $v_{n}$. The message can be given by
    \begin{equation}\label{eq3}
      \bm{s}^{(g)}_{n,m}= \varphi^{\rm M, (\it g)}(\bm{s}^{(g-1)}_{n}, \bm{s}^{(g-1)}_{m},w_{n,m}),
    \end{equation}
    where $\varphi^{\rm M, (\it g)}(\cdot)$ is a message function.
\item[$\bullet$] Aggregation. According to the generated message, the node $v_{n}$ aggregates the messages of all its neighbor nodes, as
    \begin{equation}\label{eq4}
      \bm{\hat{s}}^{(g)}_{n}=\mathop{\rm Agg}\limits_{m \in{\mathcal{N}(n)}}(\bm{s}^{(g)}_{n,m}),
    \end{equation}
    where $\rm Agg(\cdot)$ is an aggregation function, such as a summation function, that can handle a variable number of input vectors.
\item[$\bullet$]Feature update. The feature of $v_{n}$ at the $g$-th layer can be updated by merging the $\bm{s}^{(g-1)}_{n}$ and $\bm{\hat{s}}^{(g)}_{n}$, as
    \begin{equation}\label{eq5}
       \bm{s}^{(g)}_{n}=\varphi^{\rm U, (\it g)}(\bm{s}^{(g-1)}_{n},\bm{\hat{s}}^{(g)}_{n}),
    \end{equation}
    where $\varphi^{\rm U, (\it g)}(\cdot)$ is an update function.
\end{enumerate}

\textbf{Prediction output.} After $G$ layers, the updated features are $\{\bm{s}^{(G)}_n\}^{N}_{n=1}$ for each graph.
For a graph-level prediction task, we need to obtain a graph-level representation that represents the global state to predict the UE position. 
Therefore, a pooling module (e.g., global average pooling) is applied to process the node features and aggregate them into a graph-level representation.
Next, the pooled features are input into a fully connected layer to output the prediction results.

Based on the above analysis, the topology-invariant design of GNNs allows them to handle changes in the number of APs. This is because each node generates a fixed-length representation through message passing and aggregation, regardless of the total number of nodes in the graph.
However, three problems remain unresolved.
First, the amplitude and phase of the CSI are affected by different sources of noise and error, with amplitude often affected by hardware imperfections and environmental interference, while phase is sensitive to factors such as carrier frequency offset and synchronization errors. These differences introduce inconsistencies in the reliability of data. Directly inputting the raw CSI into the GNN may degrade the localization performance.
Second, the bandwidth of each AP may fluctuate dynamically due to external interference, adaptive frequency selection, and changes in communication demand. These bandwidth variations result in changes in the size of the CSI, making it challenging to input the data into the GNN which requires consistent data dimensions for each node during processing.
Third, traditional meta-learning strategies are primarily designed for multi-task scenarios in areas like image recognition, where task distributions are relatively stable and well-defined. However, cross-scenario localization tasks involve dynamic task distributions and complex environmental variations. These unique characteristics are not adequately considered in existing meta-learning strategies, leading to poor adaptation.
Therefore, we propose \name to address the above problems.

\section{Overview of Meta-SimGNN}\label{Overview of Meta-SimGNN}
To address the issue of environmental change and dynamic device configurations, we propose a novel adaptive and robust localization system, namely \name.
As depicted in Fig.~\ref{Meta-SimGNN}, \name includes three modules: CSI collection, fine-grained CSI graph construction, and location estimation. The implementation of \name can be divided into two stages: meta-training and fine-tuning.

\textbf{Meta-training stage.} In this stage, the proposed \name aims to train a set of meta-parameters that can be used for fast adaptation to the new scenario.
First, we process the historical CSI collected in the existing dataset to remove noise and other error components from the CSI measurements.
After that, we utilize the amplitude and phase information to construct fine-grained CSI graphs, with the help of an amplitude-phase fusion method based on flip techniques and a feature extraction method based on spatial pyramid pooling.
The fine-grained CSI graphs and the corresponding GNN are trained using meta-learning to obtain the meta-parameters.

\textbf{Fine-tuning stage.} The meta-parameters obtained during the meta-training stage are fine-tuned in this stage for localization in a new scenario.
Initially, a few CSI samples collected are preprocessed and constructed into the fine-grained CSI graphs as in the meta-training stage. Then, \name selects the meta-parameters by comparing the similarity between the new scenario and the historical scenarios and further fine-tunes the parameters of the GNN to adapt to the new scenario.

\section{Fine-Grained CSI Graph Construction}\label{CSI Graph Construction}
In this section, we describe how to construct a fine-grained CSI graph using amplitude and phase information with three steps: 1) Convert the CSI into an image for each AP using the proposed amplitude-phase fusion method. 2) Extract dimension-consistent features adopting the spatial pyramid pooling. 3) Construct a fine-grained CSI graph, using the obtained features.

\begin{figure}[t]
    \centering
    \begin{subfigure}[b]{0.24\textwidth}
        \includegraphics[width=\textwidth]{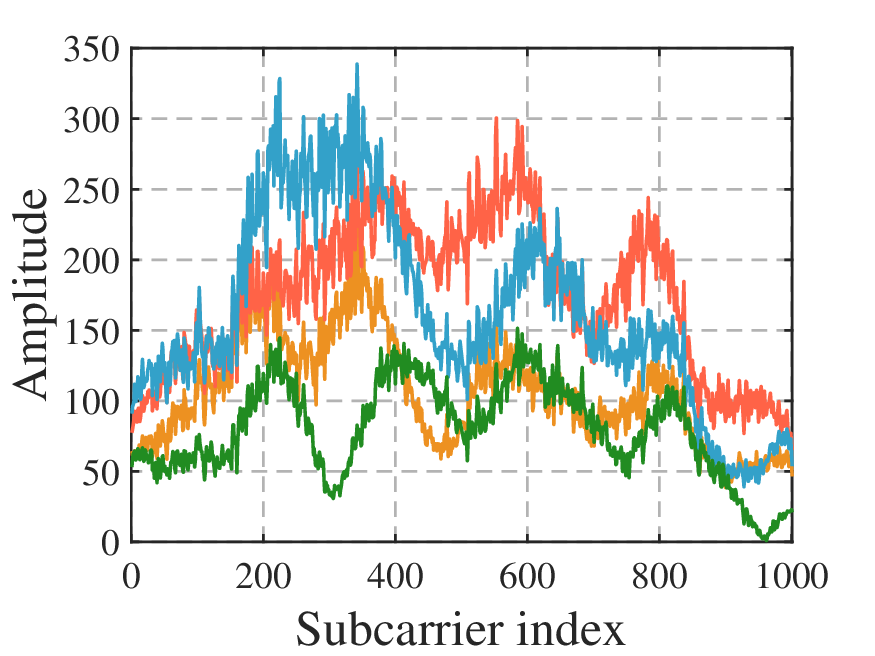}
        \caption{Before filtering.}\label{filter-csi-a}
    \end{subfigure}
    \begin{subfigure}[b]{0.24\textwidth}
        \includegraphics[width=\textwidth]{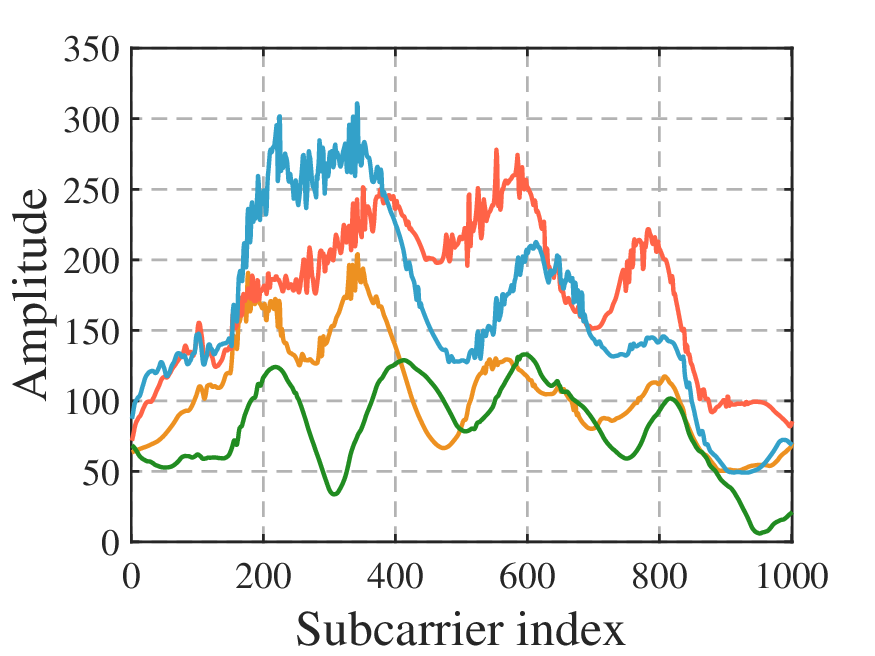}
        \caption{After filtering.}\label{filter-csi-b}
    \end{subfigure}
    \caption{CSI amplitude before and after filtering.}
    \label{filter-csi}
    \centering
    \begin{subfigure}[b]{0.24\textwidth}
        \includegraphics[width=\textwidth]{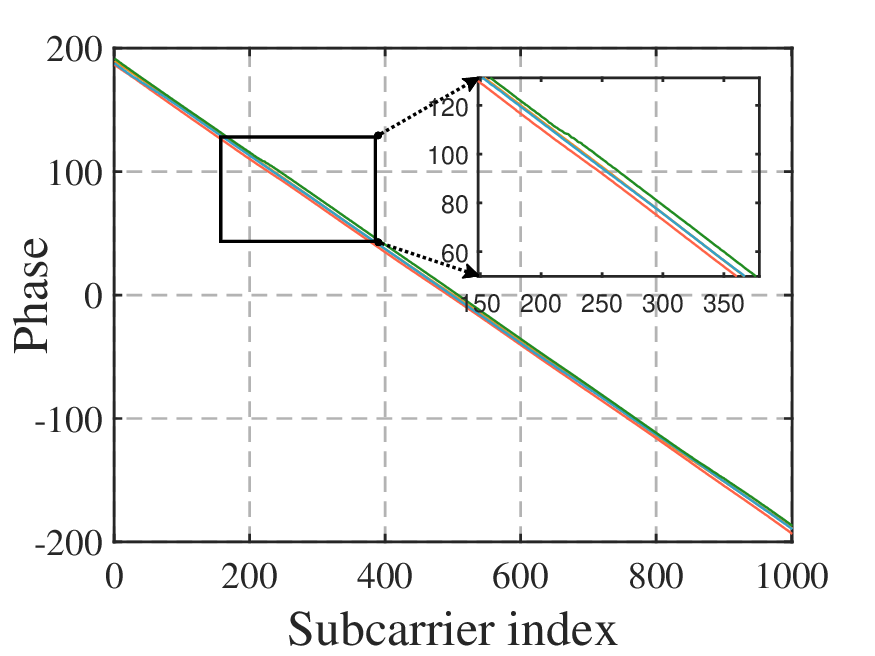}
        \caption{Before eliminating.}
        \label{modified-phase-a}
    \end{subfigure}
    \begin{subfigure}[b]{0.24\textwidth}
        \includegraphics[width=\textwidth]{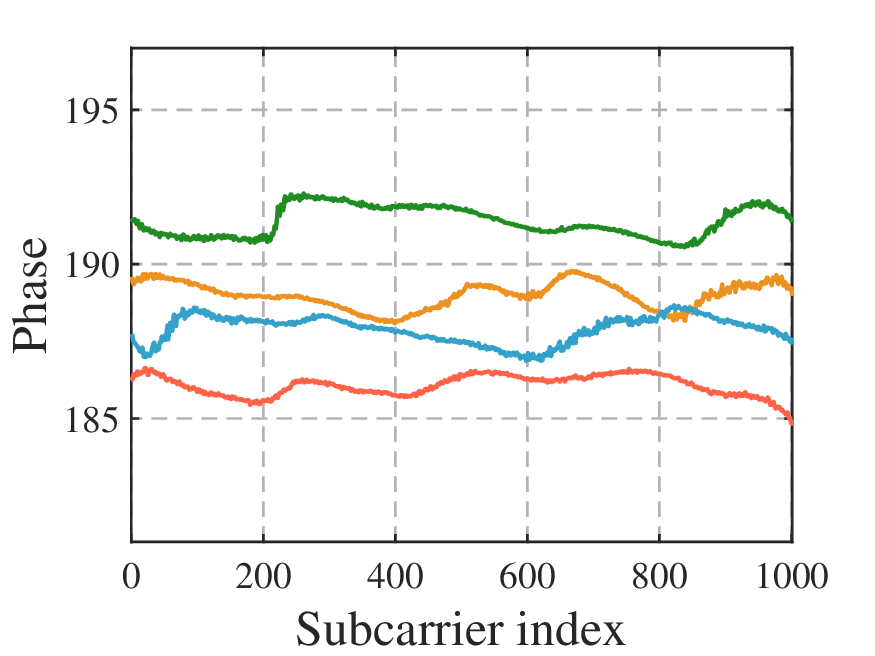}
        \caption{After eliminating.}
    \end{subfigure}
    \caption{ CSI phase before and after eliminating offsets.}
    \label{modified-phase}
    \vspace{-3ex}
\end{figure}

\subsection{Amplitude-Phase Fusion}\label{Amplitude-Phase Fusion}
Amplitude provides robust and stable signal strength characteristics and is relatively insensitive to noise, but it offers limited spatial resolution. Phase, on the other hand, captures fine-grained geometric and multipath information, but it is sensitive to hardware imperfections and noise. To balance these trade-offs, we propose an amplitude-phase fusion method to improve localization robustness.

Before utilizing the collected CSI for amplitude-phase fusion, it is necessary to clean the data to reduce the noise and the error from the hardware.
For the amplitude over different subcarriers in Fig.~\ref{filter-csi-a}, we can clearly observe the existence of noise. 
To address it, the filtering algorithm~\cite{fang2016channel,liu2015contactless} based on the discrete wavelet transform can be used to achieve smooth denoising while well preserving the signal characteristics, as shown in Fig.~\ref{filter-csi-b}.
As shown in Fig.~\ref{modified-phase}, for the phase over different subcarriers, there are errors caused by the carrier frequency offset and sampling time offset, 
and existing sanitization algorithms~\cite{kotaru2015spotfi,liu2021fire,ratnam2024optimal} can be utilized to eliminate those offsets.
After processing, we aim to input the CSI into the GNN for localization. According to Section~\ref{CSI Model}, each CSI sample consists of four dimensions: the transmit antenna dimension, the receive antenna dimension, the subcarrier dimension, and the dimension representing the complex values constructed from amplitude and phase.
Thus, it is not suitable for direct usage since most existing neural networks cannot process such a kind of data structure. To address this issue, we aim to fuse each CSI sample into one CSI image due to the success of image tasks using neural networks~\cite{egmont2002image}. Existing work~\cite{gao2023metaloc} used CSI of three WiFi APs to construct one image, but this method faces the challenge of variations in the number of APs. Thus, we opt to convert the CSI to an image for each AP.

CSI consists of both amplitude and phase components, and they can serve as two distinct channels within the RGB channels of CSI. Specifically, for the amplitude/phase channels, the size of the CSI is $N^{\rm tx}_{n} \times N^{\rm rx} \times K_{n}$, and we need to reshape it into a two-dimensional matrix with the size being $N^{\rm img}_{n} \times N^{\rm img}_{n}$, where $N^{\rm img}_{n}=\lceil \sqrt{N^{\rm tx}_{n} \times N^{\rm rx} \times K_{n}} \rceil$ and $\lceil \cdot \rceil$ is a ceiling function.
Here, we choose a reshaped dimension close to the CSI size since it can maximize the integrity of the CSI and avoid inserting too much invalid information.
Moreover, a square matrix/image is a common form used in the input images of CNN, facilitating convolution and pooling.

For the remaining channel, we can choose to either replicate the amplitude or replicate the phase. Note that the acquired phase information is susceptible to carrier frequency offset and sampling frequency offset, while the amplitude is less affected.
Thus, we opt to use the amplitude channel as the third one.
Moreover, to improve the robustness, we further apply the up-down flipping to realize data augmentation, since it changes the spatial arrangement of the subcarriers, effectively simulating variations in signal propagation patterns.

\begin{figure}[t]
    \centering
    \begin{subfigure}[b]{0.1\textwidth}
        \includegraphics[width=\textwidth]{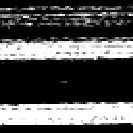}
        \caption{R.}
    \end{subfigure}
    \hspace{0.006\textwidth}
    \begin{subfigure}[b]{0.1\textwidth}
        \includegraphics[width=\textwidth]{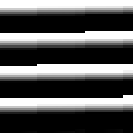}
        \caption{B.}
    \end{subfigure}
        \hspace{0.006\textwidth}
    \begin{subfigure}[b]{0.1\textwidth}
        \includegraphics[width=\textwidth]{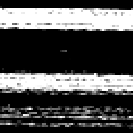}
        \caption{G.}
    \end{subfigure}
    \hspace{0.006\textwidth}
    \begin{subfigure}[b]{0.1\textwidth}
        \includegraphics[width=\textwidth]{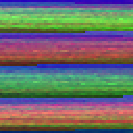}
        \caption{RGB}
    \end{subfigure}
    \vspace{-1ex}
    \caption{The CSI image for different channels at location 1.}
    \label{AP1_RP1_0}
    \centering
    \begin{subfigure}[b]{0.1\textwidth}
        \includegraphics[width=\textwidth]{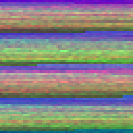}
        \caption{Loc.~1.}
    \end{subfigure}
    \hspace{0.006\textwidth}
    \begin{subfigure}[b]{0.1\textwidth}
        \includegraphics[width=\textwidth]{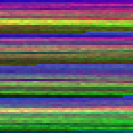}
        \caption{Loc.~2.}
    \end{subfigure}
    \hspace{0.006\textwidth}
    \begin{subfigure}[b]{0.1\textwidth}
        \includegraphics[width=\textwidth]{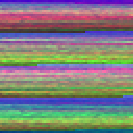}
        \caption{Loc.~3.}
    \end{subfigure}
    \hspace{0.006\textwidth}
    \begin{subfigure}[b]{0.1\textwidth}
        \includegraphics[width=\textwidth]{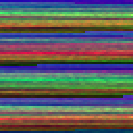}
        \caption{Loc.~4.}
    \end{subfigure}
    \vspace{-1ex}
    \caption{CSI images measured from four different locations.}
    \label{AP1_RGB}
    \vspace{-3ex}
\end{figure}

Until now, we can show the constructed CSI image with RGB channels in Fig.~\ref{AP1_RP1_0}, and its size is $N^{\rm img}_{n} \times N^{\rm img}_{n} \times 3$.
The data in the R, B, and G channels are amplitude, phase, and amplitude after up-down flipping, respectively. We first convert the data of three channels into grayscale images and then fuse them into an RGB image.
Fig.~\ref{AP1_RGB} shows the CSI images measured from four different locations.
It can be seen that they are different from each other, providing unique location features for indoor localization.

\subsection{Feature Extraction}
After constructing the CSI image, we now extract features from it.
Most of the existing work adopts the CNN directly, thanks to the powerful feature mapping capabilities of the convolutional operations~\cite{li2021survey}.
However, it cannot be used directly in this paper because
dynamic AP configurations (e.g., bandwidth, number of antennas, etc.) lead to varying sizes of the constructed images. If the same CNN network is applied, the output feature dimensions differ for different APs. An intuitive solution is to use different CNNs. However, this results in higher training complexity, and feature output consistency among different CNNs is difficult to guarantee.

\begin{figure}[t]
  \centering
  \includegraphics[width=0.47\textwidth]{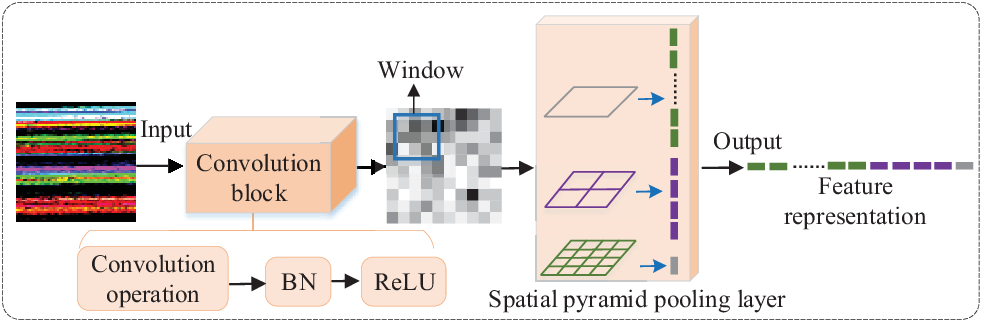}
  \caption{Feature extraction based on spatial pyramid pooling.}
  \label{CNN_SPP_extract}
  \vspace{-3ex}
\end{figure}

To tackle this problem, we adopt the spatial pyramid pooling~\cite{he2015spatial} to design a feature extraction method.
As shown in Fig.~\ref{CNN_SPP_extract},
it consists of convolutional layers with rectified linear units (ReLUs) and batch normalization (BN) layers~\cite{jiao2023enhancing}, a spatial pyramid pooling layer, and a fully connected layer.
Specifically, the generated CSI image is first translated into a feature map with the size being $z\times z$, and note that $z$ is related to the number of subcarriers $K_n$.

Then, the CSI feature map is input into the spatial pyramid pooling layer for down-sampling.
It divides the CSI feature map into grid areas of multiple scales (or pyramid levels), performs pooling operations on each area to capture global and local multi-scale information, and generates fixed-length feature representations.
To be specific, we consider an $L^{\rm pool}$-level pyramid. The output feature size of the $l$-level pyramid is  $c_{l} \times c_{l}$.
To obtain $c_{l} \times c_{l}$ features, we perform sliding window pooling of the CSI feature map at each pyramid level, where the window size is  $\lceil z/c_{l} \rceil$ and stride size is $\lfloor z/c_{l} \rfloor$ with $\lceil \cdot \rceil$ and $\lfloor \cdot \rfloor$ denoting ceiling and floor operations, respectively.
After completing the pyramid pooling, the fully connected layer concatenates $L^{\rm pool}$ outputs to obtain a fixed-length feature representation.
This pooling part reduces the size of the CSI feature map and extracts important features, which helps to reduce the computational complexity~\cite{hoang2020cnn} and prevent overfitting.

\subsection{Fine-Grained CSI Graph Construction} \label{sec:csi_graph}
\begin{figure}[t]
  \centering
  \vspace{-1ex}
  \includegraphics[width=0.48\textwidth]{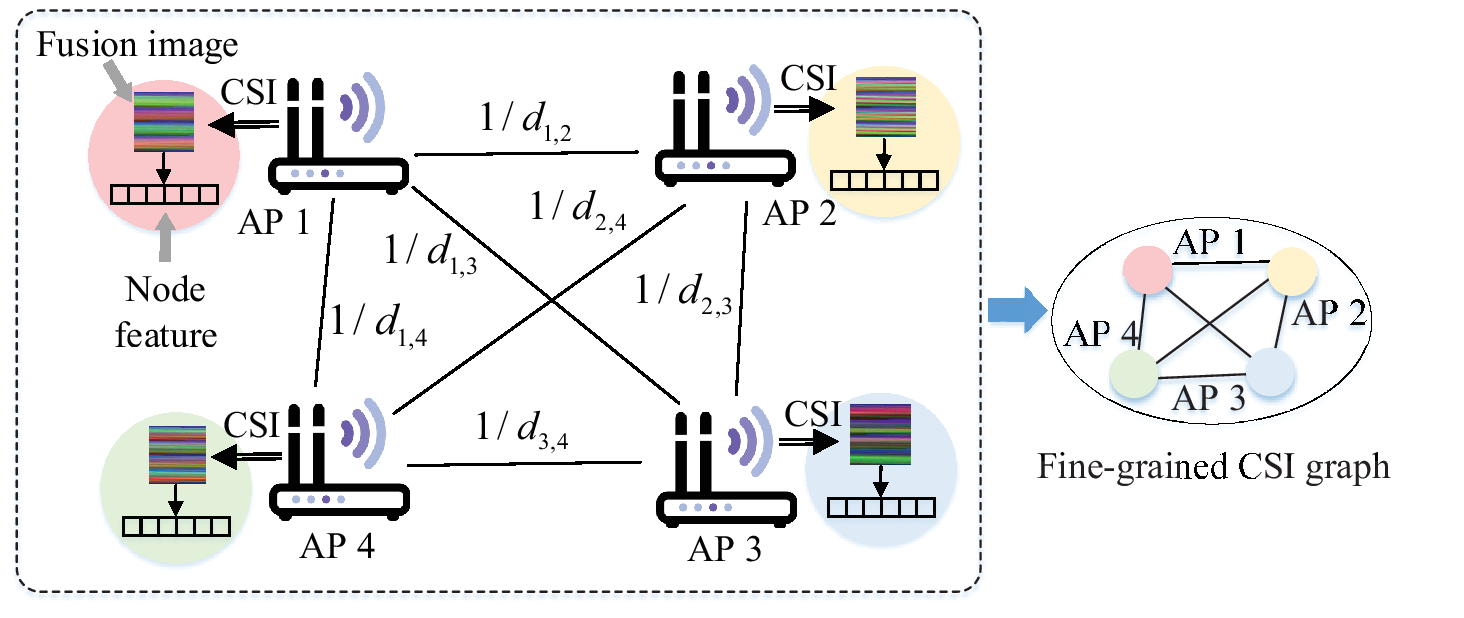}
  \vspace{-1ex}
  \caption{The fine-grained CSI graph construction method.}
  \label{CSI_fine_grained_graph}
  \vspace{-1ex}
\end{figure}

Until now, we have obtained the feature for each AP, and
we now aim to use features from all APs to construct a graph in this section, which can adapt to the varying number of APs.
To this end, a novel fine-grained CSI graph construction is proposed, as shown in Fig.~\ref{CSI_fine_grained_graph}.
Specifically, as introduced in Section~\ref{Graph Neural Network}, a graph can be represented as $\mathcal{G} = (\mathcal{V}, \mathcal{E}, \mathcal{X})$, which requires node features and edge features.
For $N$ WiFi APs, each AP is regarded as a graph node, and the CSI feature, denoted by $\bm{x}^{(0)}_{n}$, obtained after amplitude-phase fusion and feature extraction, is used as the graph node features.

Due to the limited indoor space and the small scale of AP deployment, it is reasonable to assume that any two APs in the indoor scenario are connected in the graph, and the constructed graph is a complete one. To capture the spatial relationship between any two APs and signal propagation characteristics, the weights in the adjacency matrix $\bm{A} \in \mathbb{R}^{N \times N}$ should be related to the distance between any two APs. Since each AP node would update its feature using the features from its neighbor and features from far away APs should be naturally suppressed, we consider that the weight is set as the inverse of the distance between APs, and the corresponding adjacency matrix is given by
\begin{equation}\label{eq7}
  \bm{A}_{n,m}=
  \left \{
  \begin{aligned}
    &1/d_{n,m},~{\rm{if}}~n\in\mathcal{N}(v), \\
    &0,~\rm otherwise,
  \end{aligned}
  \right.
\end{equation}
where $d_{n,m}$ is the Euclidean distance between the $n$-th AP and the $m$-th AP.
With equation~\eqref{eq7},
the spatial perception of the graph is effectively improved.

The constructed fine-grained CSI graph is further input into the GNN for location estimation. Here, we apply three graph convolutional layers to capture the relationships between nodes through a feature propagation mechanism.
At every layer, the feature of each node is updated by aggregating information from its neighboring nodes through graph convolution operations. Each graph convolutional layer is followed by a ReLU activation and a BN layer to enhance non-linearity and stabilize training.
After the three graph convolutional layers, we employ a global average pooling layer to aggregate the node features into a graph-level representation.
This representation is then passed through a fully connected layer to perform accurate location estimation.
Notably, the GNN's insensitivity to the number of nodes makes it robust against variations in the number of APs.
To train the model, we use the mean square error (MSE) loss function since the localization task can be regarded as a regression problem.
Our optimization objective is to minimize the MSE between the predicted UE location and the ground-truth location. The MSE loss function can be expressed as
\begin{equation}\label{MSE}
	\mathcal{L}(f_{\bm{\theta}})=\sum\limits_{\bm{x}_{q},y_{q} \sim \mathcal{T}_{p}} \| f_{\bm{\theta}}(\bm{x}_{q})-y_{q} \|^{2}_{2},
\end{equation}
where $\bm{x}_{q}$ denotes the CSI features of the $q$-th UE, which serve as input to the GNN model, $f_{\bm{\theta}}(\bm{x}_{q})$ denotes the predicted location of the $q$-th UE, and $y_{q}$ is the ground-truth location of the $q$-th UE.

\section{Similarity-Guided Meta-Learning}
\label{ML-GDS}
\begin{figure*}[!t]
  \centering
  \includegraphics[width=0.8\textwidth]{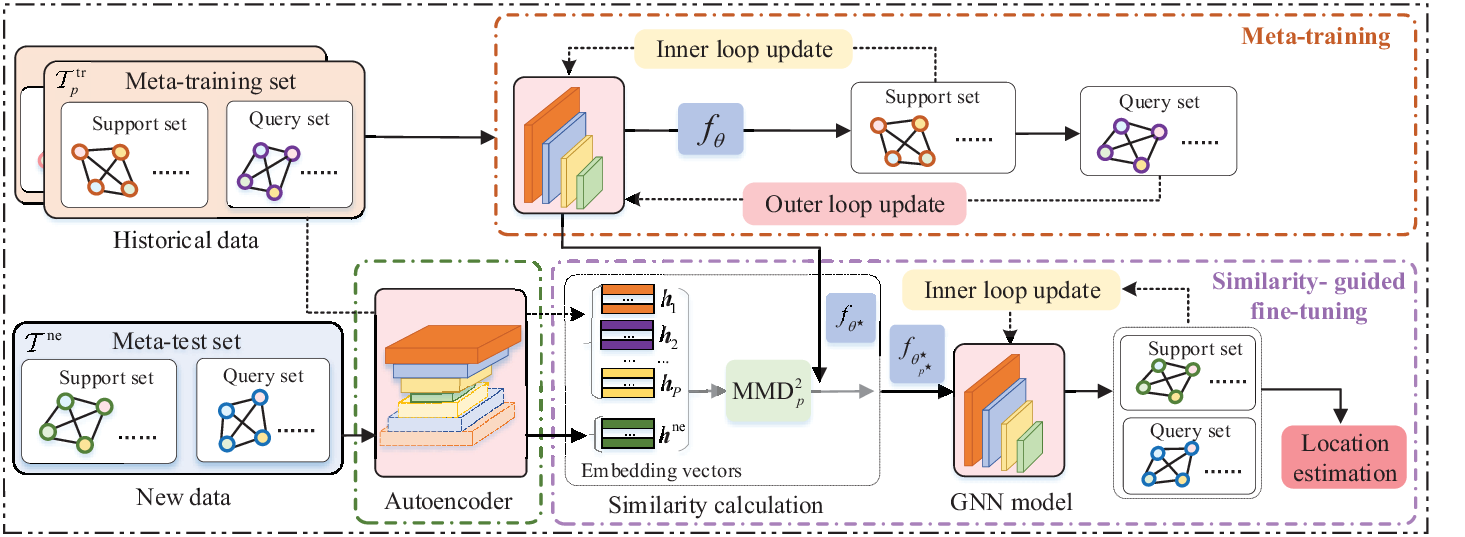}\\ 
  \caption{The proposed similarity-guided meta-learning strategy.}
  \label{Meta-SimGNN_network}
  \vspace{-2ex}
\end{figure*}

In the previous section, we have constructed a fine-grained CSI graph and the corresponding GNN, which can adapt to the dynamic device configuration in the same scenario.
However, when deploying the localization network in new scenarios, we still need to re-collect data and re-train the model, resulting in high overhead.
To address this problem, we aim to employ the meta-learning. Nevertheless, unlike existing meta-learning approaches designed to handle diverse visual tasks, our approach targets different scenarios. 
Specifically, unlike original strategies that train meta-parameters for all scenarios, we train scenario-specific meta-parameters for each historical scenario.
Then, during the meta-testing stage, we select the scenario-specific parameters that are most similar to the new scenario for fine-tuning, thereby achieving fast adaptive localization.
In the localization task, there are typical scenarios such as offices, large shopping malls, and classrooms. New scenarios are often slight variations of these typical ones. If we can leverage parameters trained on similar typical scenarios, these parameters can quickly adapt to new scenarios. Based on this idea, we propose a similarity-guided meta-learning strategy, as shown in Fig.~\ref{Meta-SimGNN_network}.
It consists of two stages: meta-training and similarity-guided fine-tuning.
The first stage trains scenario-specific meta-parameters for each historical scenario using the collected CSI dataset, and the second stage first selects the historical scenario that is most similar to the new scenario and then fine-tunes the parameters using the limited CSI from the new scenario.
For ease of expression, the GNN established in the previous section is denoted by $f_{\bm{\theta}}$ with $\bm{\theta}$ being the trainable parameters.

\subsection{Meta-Training}
Considering there are datasets $\{\mathcal{T}^{\mathrm{tr}}_{p},p=1,\cdots,P\}$ collected from $P$ historical scenarios, we aim to train model parameters for each scenario. To obtain well-suited model parameters that can represent one typical scenario, we apply a traditional meta-training strategy (i.e., the MAML strategy~\cite{finn2017model}) for each scenario. Specifically, the dataset from the $p$-th scenario is randomly divided into $I_p$ groups, with each group further split into a support set $\mathcal{S}_{p,i}$ and a query set $\mathcal{Q}_{p,i}$. The training process is divided into two parts: an inner loop and an outer loop.
The inner loop is to update the group-specific parameters using the support set, while the outer loop first selects $I^{\mathrm{tr}}_p$ groups and then updates the parameters using the query set and the group-specific parameters of these groups.

In the inner loop, we use the gradient descent update to obtain the group-specific parameters $\bm{\theta}'_{p,i}$, which is given by
\begin{equation}\label{eq8}
   \bm{\theta}'_{p,i} \leftarrow \bm{\theta}-\alpha \nabla_{\bm{\theta}}\mathcal{L}(f_{\bm{\theta}}; \mathcal{S}_{p,i}),
\end{equation}
where $\alpha$ is the learning rate in the inner loop and $\mathcal{L}(f_{\bm{\theta}}; \mathcal{S}_{p,i})$ is the loss over the support set $\mathcal{S}_{p,i}$.
In the outer loop, the loss over the query sets of $I^{\mathrm{tr}}_p$ groups is calculated as $\sum\limits^{I^{\mathrm{tr}}_p}_{i=1}\mathcal{L}(f_{\bm{\theta}'_{p,i}}; \mathcal{Q}_{p,i})$. To update this loss, the stochastic gradient descent (SGD) is adopted, and the updated parameters are given by
\begin{equation}\label{eq10}
  \bm{\theta} \leftarrow \bm{\theta}-\beta\nabla_{\bm{\theta}}\sum\limits^{I^{tr}_{p}}_{i=1}\mathcal{L}(f_{\bm{\theta}^{'}_{p,i}}; \mathcal{Q}_{p,i}),
\end{equation}
where $\beta$ is the learning rate in the outer loop.

Based on the above steps, the detailed meta-training procedure can be summarized in Algorithm~\ref{alg1}. By applying it, we can obtain the scenario-specific parameters for each historical scenario, denoted by $\bm{\theta}^{\star}_{p}$.

\begin{algorithm}[htbp]
\caption{Meta-training procedure}
\label{alg1}
{\normalsize
\begin{algorithmic}[1]
\STATE \textbf{Input}: CSI datasets $\mathcal{T}^{\rm tr}_{p}, p=1,2, ..., P$, inner network learning rate $\alpha$, outer network learning rate $\beta$, iteration count $L^{\rm epo}$, meta batch size $I^{\rm tr}_p$, and the number of scenario-level inner update steps $L^{\rm in}$.
\STATE \textbf{Initialization}: Randomly initialize $\bm{\theta}$;
\FOR {$p = 1 : P$}
\FOR {$l = 1 : L^{\rm epo}$}
\STATE Sample a batch of CSI samples $\{\mathcal{T}^{\rm tr}_{p,i}\}^{I^{\rm tr}_p}_{i=1}$;
\FOR {$i = 1 : I^{\rm tr}_p$}
\STATE Update parameters $\bm{\theta}'_{p,i}$ using equation~\eqref{eq8} with the number of iteration being $L^{\rm in}$;
\ENDFOR
\STATE $\bm{\theta} \leftarrow \bm{\theta}-\beta \nabla_{\bm{\theta}}\sum\limits^{I^{\rm tr}_p}_{i=1} \mathcal{L}(f_{\bm{\theta}'_{p,i}}; \mathcal{Q}_{p,i})$;
\ENDFOR
\STATE $\bm{\theta}^{\star}_{p} \leftarrow \bm{\theta}$.
\ENDFOR
\end{algorithmic}}
\end{algorithm}

\subsection{Similarity-Guided Fine-Tuning}
In the similarity-guided fine-tuning stage, the collected CSI dataset $\mathcal{T}^{\rm ne}$ of the new scenario is divided into a new support set $\mathcal{S}^{\rm ne}$ and a new query set $\mathcal{Q}^{\rm ne}$.
We need to first select the historical scenario that is most similar to the new one, and then the corresponding parameters are fine-tuned using the $\mathcal{S}^{\rm ne}$.
We use the datasets of the historical scenarios and the new scenario to calculate the similarity.
To measure the similarity, it is hard to directly use the original collected CSI sample due to its relatively large size and the presence of some other information. Thus, we first utilize an autoencoder~\cite{ding2019deep} to compress the CSI to obtain the high-level representation for the scenario. Specifically, we also adopt the graph construction in Section~\ref{sec:csi_graph} and use the CSI graph as the input.\footnote{We assume that the collected dataset for training is measured under the unified device configuration, and thus, we do not need the feature extraction here.} The autoencoder consists of an encoder and a decoder, where the encoder aims to compress the CSI graph into the low-dimensional vector and the decoder aims to recover the original CSI graph. The encoder has three graph convolutional layers (each followed by a ReLU function and a dropout layer) and a pooling layer. The size of features decreases with the feature propagation and the features become a vector, i.e., the representation of the scenario, after the pooling layer. The decoder also has three graph convolutional layers for increasing the size of the feature. To ensure the effectiveness of compression, the autoencoder is trained using all collected datasets from $P$ historical scenarios and the MSE between the input CSI graph and output CSI graph is adopted as the loss function.

After using the encoder in the autoencoder to compress the CSI graph, we now only need to measure the scenario similarity. Let $\bm{h}_{p,i}$ denote the representation of the $p$-th scenario compressed from the $i$-th CSI graph and the representation of the new scenario is denoted by $\bm{h}_{i}^{\mathrm{ne}}$. Then, we can use maximum mean discrepancy (MMD)~\cite{gretton2012kernel} to measure the similarity between $\{\bm{h}_{p,i},\forall i\}$ and $\{\bm{h}_{i}^{\mathrm{ne}},\forall i\}$, which is a widely used method for comparing whether two probability distributions are the same. Specifically, it measures the distance between two distributions by comparing the difference between the averages of samples from these distributions in a certain feature space. Let $\phi(\cdot)$ denote the feature mapping function, and the MMD between the $p$-th scenario and the new scenario can be expressed as
\begin{align}\label{eq:mmd}
  {\rm MMD}^{2}_p=&\left|\left| \frac{1}{I^{\mathrm{sa}}_p}\sum\limits^{I^{\mathrm{sa}}_p}_{i=1}\phi(\bm{h}_{p,i}) - \frac{1}{I^{\mathrm{sa,n}}}\sum\limits^{I^{\mathrm{sa,n}}}_{i=1}\phi(\bm{h}_{i}^{\mathrm{ne}})  \right|\right|^2 \nonumber \\
  =&  \frac{1}{(I^{\mathrm{sa}}_p)^2} \!\sum_{i,j=1}^{I^{\mathrm{sa}}_p}\! \kappa (\bm{h}_{p,i}, \bm{h}_{p,j} ) \!+\! \frac{1}{(I^{\mathrm{sa,n}})^2} \!\sum_{i,j=1}^{I^{\mathrm{sa,n}}}\! \kappa (\bm{h}_{i}^{\mathrm{ne}}, \bm{h}_{j}^{\mathrm{ne}}) \nonumber \\
  &- \frac{1}{I^{\mathrm{sa}}_p I^{\mathrm{sa,n}} } \!\sum_{i,j=1}^{I^{\mathrm{sa}}_p, I^{\mathrm{sa,n}}} \!\kappa (\bm{h}_{p,i}, \bm{h}_{j}^{\mathrm{ne}}),
\end{align}
where $\kappa(\cdot,\cdot)$ is the kernel function for $\phi(\cdot)$, and $I^{\mathrm{sa}}_p$ and $I^{\mathrm{sa,n}}$ are the numbers of CSI graph samples for the $p$-the scenario and the new scenario, respectively. Here, we adopt the Gaussian kernel~\cite{sun2024mmd}, which is one of the widely used and can greatly simplify the calculation. From equation~\eqref{eq:mmd}, it can be observed that the MMD indicates the degree of difference between distributions. The smaller the MMD is, the smaller the degree of difference would be. Following this, we can select the optimal historical scenario by
\begin{equation}
    p^{\star} = \mathop{\arg\min}_{p \in \{1, 2, ..., P\}}   {\rm MMD}^{2}_{p}.
\end{equation}

Then, the support set $\mathcal{S}^{\rm ne}$ is used to adaptively update the scenario-specific meta-parameters through $L^{\rm f}$ rounds of gradient descent to fine-tune the model, which can be given as
\begin{equation}\label{eq14}
  \bm{\theta}_{E}  \leftarrow  \bm{\theta}^{\star}_{p^{\star}}-\alpha \nabla_{\bm{\theta}^{\star}_{p^{\star}}}  \mathcal{L}(f_{\bm{\theta}^{\star}_{p^{\star}}} ; \mathcal{S}^{\rm ne}).
\end{equation}
The localization performance of the fine-tuned model is then evaluated on $\mathcal{Q}^{\rm ne}$. The detailed steps of the similarity-guided fine-tuning process are summarized in Algorithm~\ref{alg2}.

\begin{algorithm}[t]
\caption{Similarity-guided fine-tuning procedure}
\label{alg2}
{\normalsize
\begin{algorithmic}[1]
\STATE \textbf{Input}: CSI dataset from the new scenario, $\mathcal{T}^{\rm ne}=[\mathcal{S}^{\rm ne}, \mathcal{Q}^{\rm ne}]$, inner network learning rate $\alpha$, and the number of update steps for fine-tuning $L^{\rm f}$.
\FOR{$p = 1: P$}
\STATE Calculate the $\mathrm{MMD}^{2}_{p}$ of the $p$-th historical scenario and the new scenario using equation~\eqref{eq:mmd};
\ENDFOR
\STATE Choose the scenario-specific meta-parameters $\bm{\theta}^{\star}_{p^{\star}}$, where $p^{\star} = \mathop{\arg\min}_{p \in \{1, 2, ..., P\}}{\rm MMD}^{2}_{p}$;
\STATE Fine-tune parameters $\bm{\theta}_{E}$ using equation~\eqref{eq14} with the number of iteration being $L^{\rm f}$.
\end{algorithmic}}
\end{algorithm}

\section{Experimental Setup}\label{Experimental Setup}
In this section, we first elaborate on \name's implementation and experimental setup, and then introduce the baseline methods used for comparison.

\subsection{Implementation and Setup}
\textbf{Implementation}: Four WiFi APs (i.e., TP-LINK AX5400 routers) are deployed to transmit WiFi signals. A computer with a two-antenna Intel AX210 network interface card (NIC) with the antenna spacing being 6~\!cm acts as a UE.
To simulate a real-world scenario, the receiver antenna height is set to 1.55~\!m, and the antenna height of the WiFi AP is 1.8~\!m.
Both the WiFi APs and Intel AX210 NIC work in the IEEE 802.11ax (WiFi 6) standard with the center frequency being 5.25~\!GHz and bandwidth randomly distributed in the set of [20, 40, 80]~\!MHz. To collect the CSI at the UE, we set the UE to work in the monitor mode and utilize the PicoScenes tool~\cite{jiang2021eliminating} with a sampling frequency of 100~\!Hz. Moreover, the software framework of the \name is implemented using Python~3.9.0 and PyTorch~1.11.0.

\begin{figure*}[t] 
    \centering
    \begin{subfigure}{0.21\textwidth}
        \centering
         \vspace{-2mm}
        \includegraphics[width=1\linewidth]{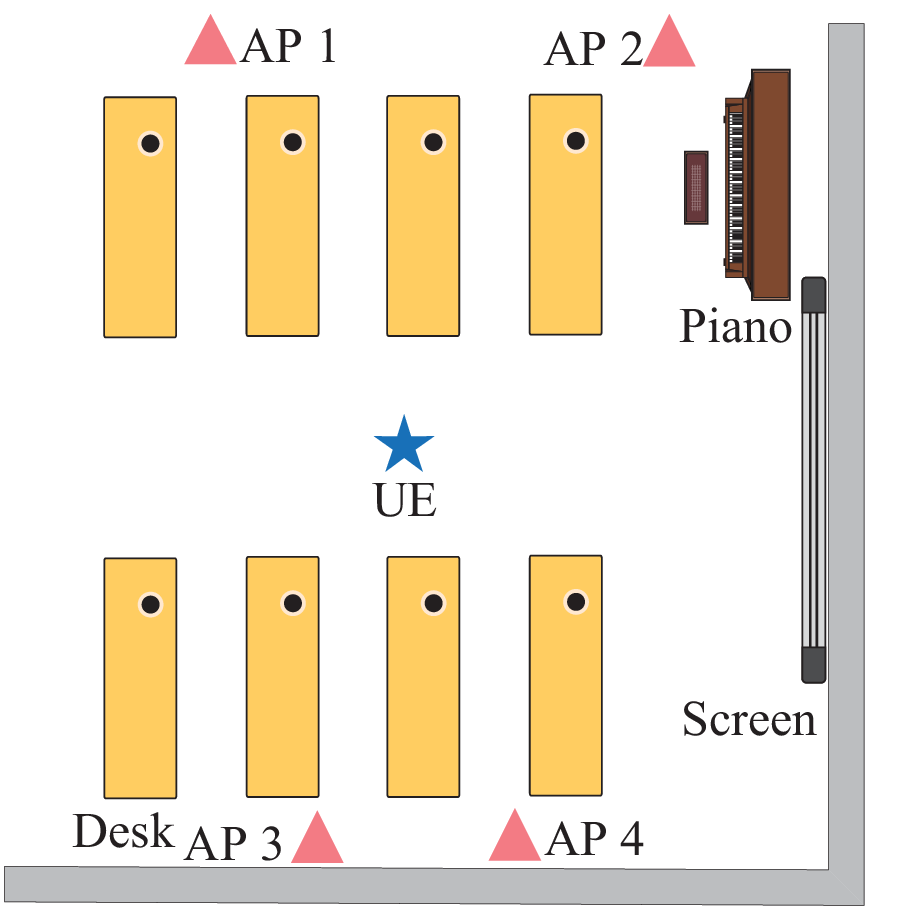}
        \caption{Scen. 1.}
        \label{fig:scen1}
    \end{subfigure}
    \hspace{-0.02\textwidth}
    \begin{subfigure}{0.21\textwidth}
        \centering
        \includegraphics[width=1\linewidth]{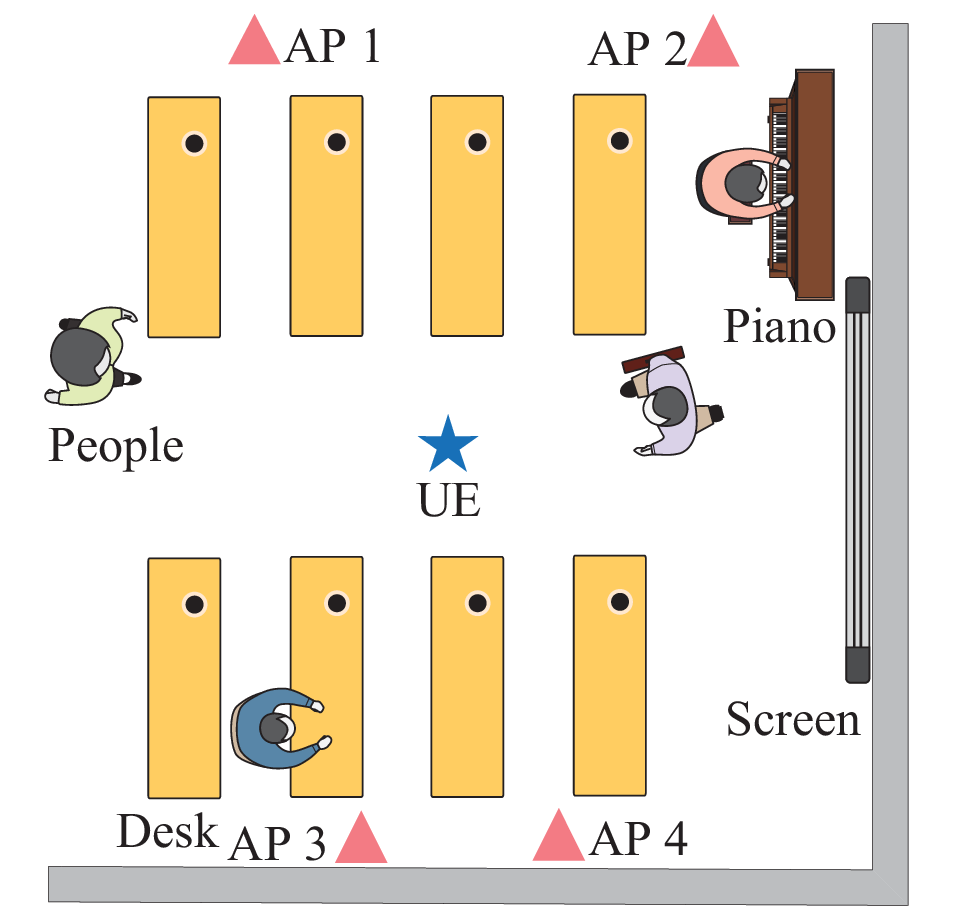}
        \caption{Scen. 1 with passersby.}
        \label{fig:scen1_pass}
    \end{subfigure}
    \hspace{-0.02\textwidth}
    \begin{subfigure}{0.21\textwidth}
        \centering
        \vspace{1mm}
        \includegraphics[width=0.89\linewidth]{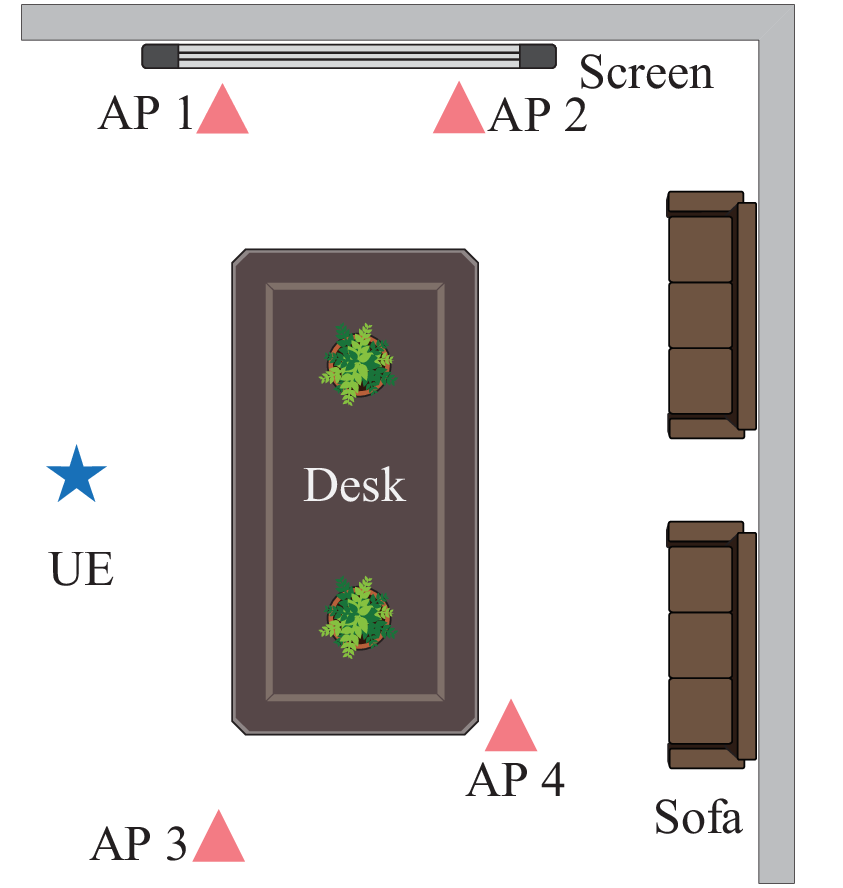}
        \caption{Scen. 2.}
    \end{subfigure}
    \hspace{-0.02\textwidth}
    \begin{subfigure}{0.21\textwidth}
        \centering
        \vspace{1.5mm}
        \includegraphics[width=0.92\linewidth]{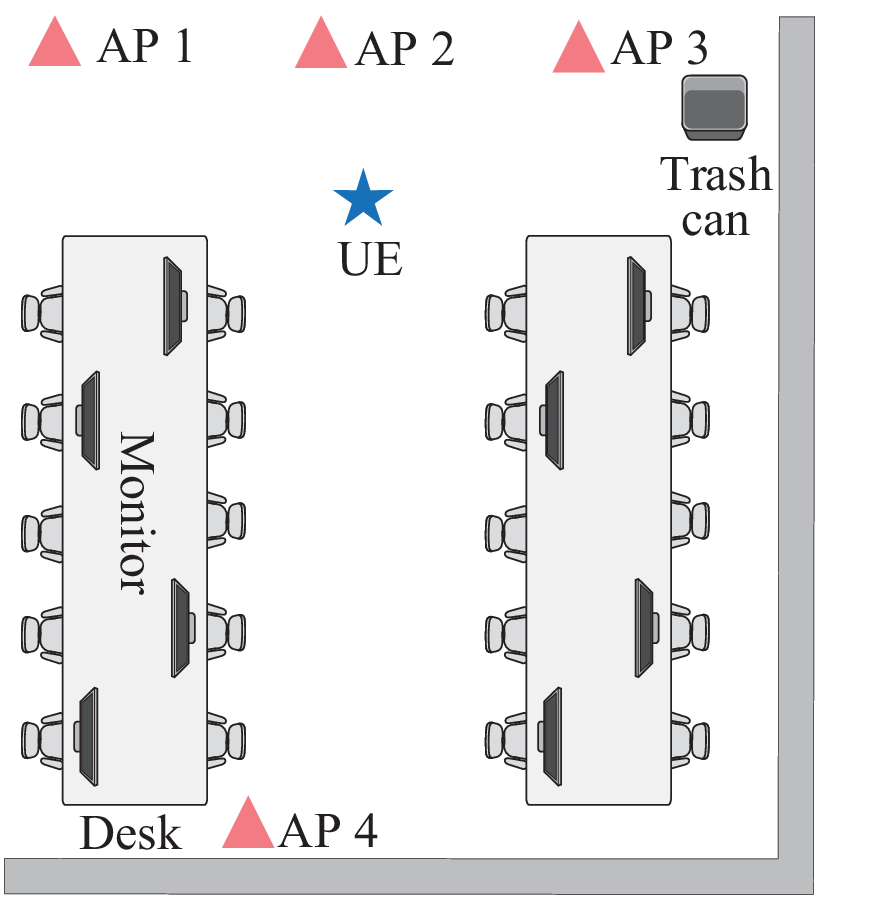}
        \caption{Scen. 3.}
    \end{subfigure}
    \hspace{-0.02\textwidth}
    \begin{subfigure}{0.18\textwidth}
        \centering
        \vspace{0.5mm} 
        \includegraphics[width=0.73\linewidth]{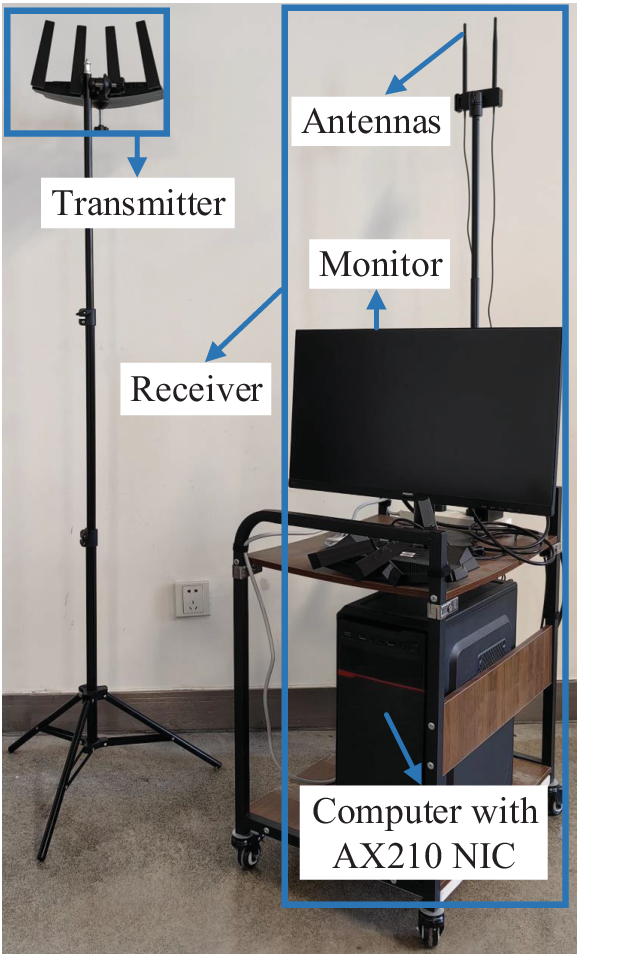}
        \caption{Prototype.}
    \end{subfigure}
    \caption{Experimental setup in three scenarios and prototype of \name.}
    \label{scenarios}
    \vspace{-3ex}
\end{figure*}

\textbf{Experiment Setup}:
We deploy WiFi APs and UE in three different scenarios: classroom, meeting room, and laboratory, as detailed in the following.
\begin{enumerate}
\item[$\bullet$] \textbf{Scenario 1}: As shown in Fig.~\ref{scenarios}(a), Scenario 1 is in an $8\times 9$~\!m classroom, where four WiFi APs are deployed at different locations to transmit signals. The UE collects CSI at each RP, and there are no obstacles interfering with the CSI collection. To increase the diversity of the CSI samples, we extended Scenario 1, as shown in Fig.~\ref{scenarios}(b), where obstacles such as people walking around interfere with the CSI collection.
\item[$\bullet$] \textbf{Scenario 2}: Scenario 2 is in an $8\times 7$~\!m meeting room, where four WiFi APs are deployed. There are no obstacles interfering with the UE collecting CSI, as demonstrated in Fig.~\ref{scenarios}(c).
\item[$\bullet$] \textbf{Scenario 3}: As depicted in Fig.~\ref{scenarios}(d), Scenario 3 is in an $8 \times 9.5$~\!m laboratory with some electronic devices. Four WiFi APs are deployed at different locations to transmit WiFi signals. When the UE collects CSI, there are no obstacles to interfere.
\end{enumerate}
In each scenario, to collect the dataset, we set reference points (RPs) with
the distance between two RPs being 0.5~\!m.
In Scenarios 1, 2, and 3, we initially collected 1000 WiFi packets on each RP.
In Scenario 1, two datasets were collected under different conditions: one without obstacle interference, and another several days later with obstacles such as people moving around. Fifteen days later, an additional 1500 WiFi packets were collected at each RP in all three scenarios.
In meta-learning, we adopt some scenarios as the historical ones with 80 CSI samples of each RP used for training, and the remaining scenario(s) as the new one(s) with 20 CSI samples used for fine-tuning. The remaining CSI samples are used for testing the performance of \name.

\subsection{Baseline Methods}\label{Baseline}

Here, we consider the following baseline methods.

$\bullet$ GNN: A GNN model with three graph convolutional layers is adopted. The model is trained using the historical CSI samples without adopting the meta-learning framework.

$\bullet$  Meta-GNN~\cite{yu2024meta}: 
Directly combine the traditional meta-learning framework (i.e., MAML) with GNN. In the meta-testing stage, the meta-parameters learned across all training scenarios serve as the initial parameters for fine-tuning.

$\bullet$ MetaLoc~\cite{gao2023metaloc}: We adopt both the MAML-TS and MAML-DG frameworks proposed in MetaLoc because they represent current state-of-the-art approaches for improving cross-scenario generalization in WiFi localization through meta-learning. 
MAML-TS trains environment-specific meta-parameters for each environment, and MAML-DG can utilize environment-specific meta-parameters from different environments to train a set of globally shared meta-parameters.
The CSI image construction method, CNN network structure, and neural network hyperparameters follow the same settings as those in MetaLoc.

In the following experiment evaluation section, unless otherwise specified, the results are presented as averaged localization errors.

\section{Experiment Evaluation}\label{Experiment Evaluation}
In this section, we provide the experimental verification and analysis of the proposed \name. We first show micro-benchmark studies to demonstrate
the effectiveness of \name.
Then, the generalization performance of \name is presented by comparing it to the baseline methods.
Finally, we evaluate the comprehensive performance of \name under real-world settings.

\subsection{Micro-Benchmark Studies}
In this subsection, we mainly focus on the effectiveness of the proposed fine-grained CSI graph construction scheme, and thus, we train and test the model in the same scenario.
To assess the impact of training data quantity, we vary the number of CSI samples used for training by adjusting the sampling densities.

\begin{figure}[t]
	\centering
    \vspace{-2ex}
	\begin{subfigure}[b]{0.48\textwidth}
		\includegraphics[width=0.98\textwidth]{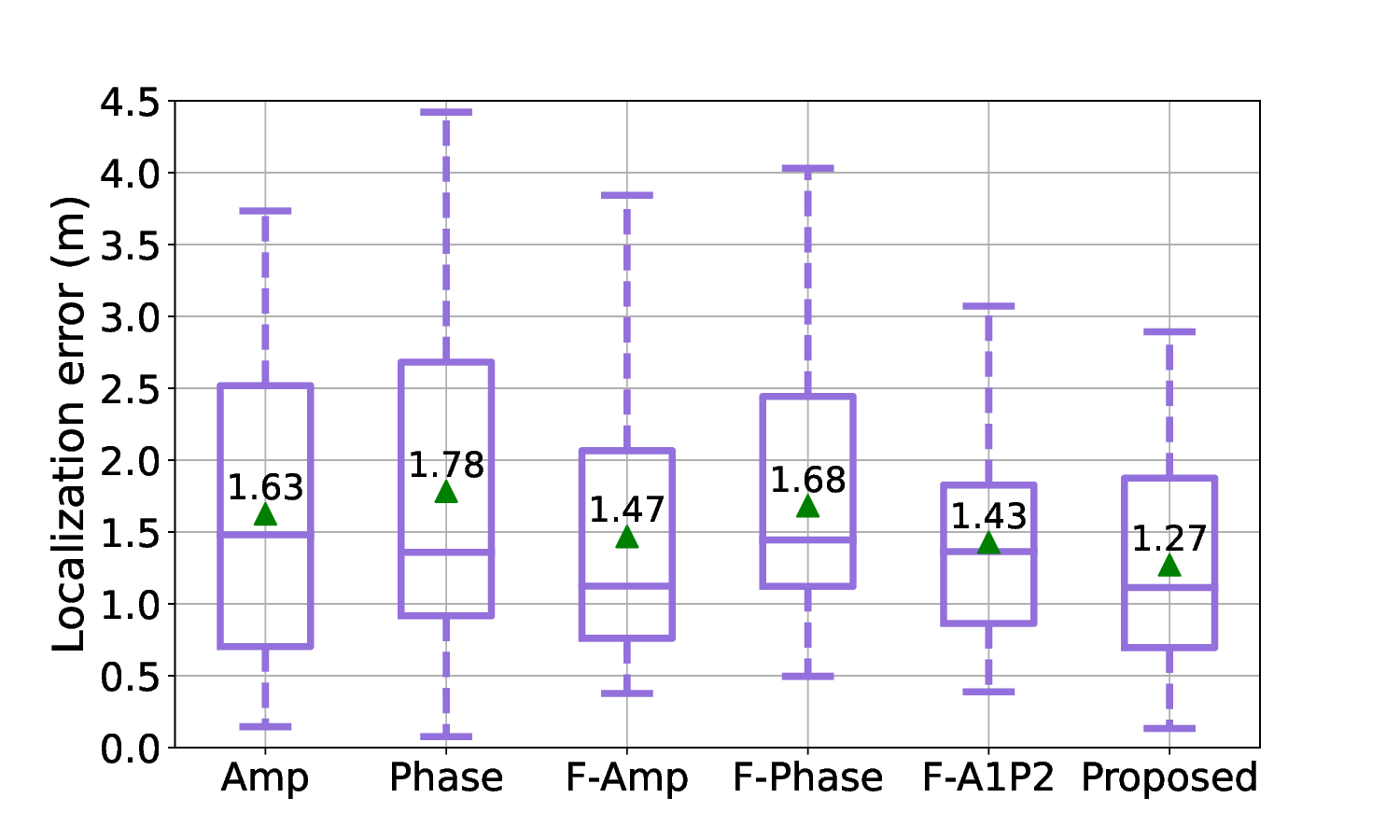}
        \vspace{-2ex}
		\caption{Scen. 1.}\label{csi_fusion_s10}
	\end{subfigure}
	\begin{subfigure}[b]{0.48\textwidth}
		\includegraphics[width=0.98\textwidth,trim=0 0 0 20,clip]{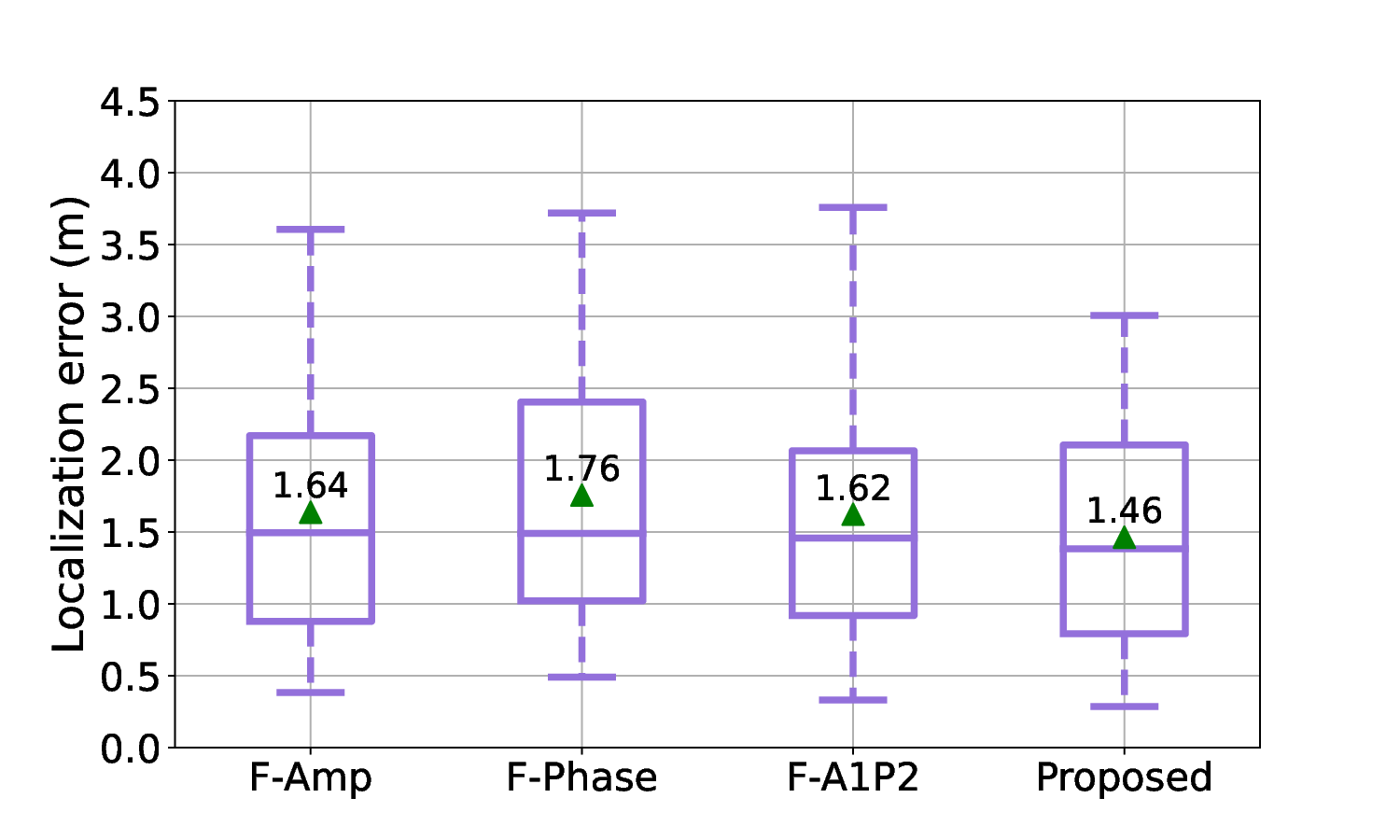}
        \vspace{-2ex}
		\caption{Scen. 1 with passersby.}\label{csi_fusion_s11}
	\end{subfigure}
	\caption{Localization performance under different CSI graphs with ``$\blacktriangle$'' representing the mean error.}
	\label{CSI_fusion}
    \vspace{-3ex}
\end{figure}

\begin{figure}[t]
	\centering
    \vspace{-2ex}
	\includegraphics[width=0.45\textwidth]{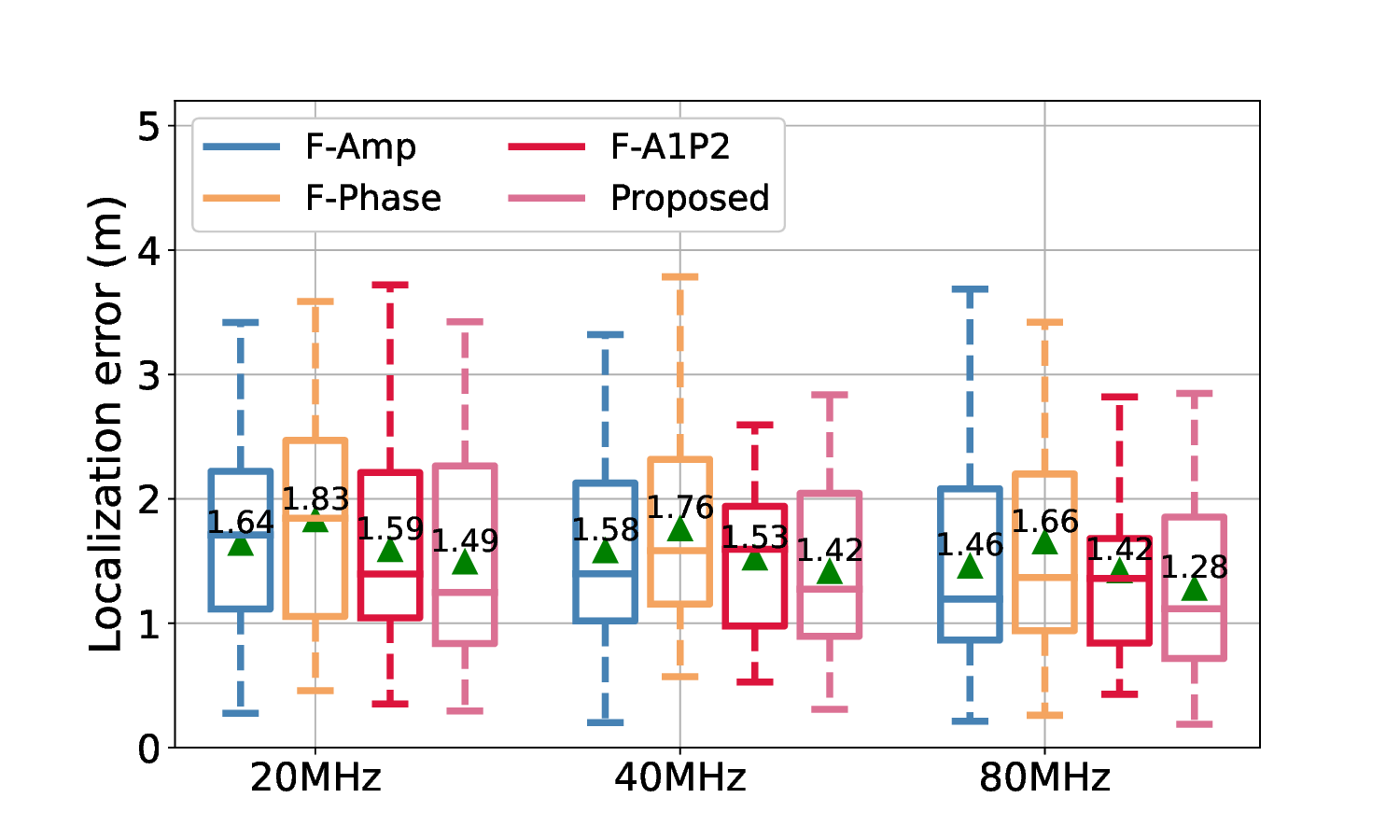}
    \vspace{-2ex}
	\caption{Localization performance under different bandwidths.}
	\label{CSI_fusion_BW}
	\vspace{-3ex}
\end{figure}
\textbf{1) Effectiveness of the constructed CSI graph.}
To demonstrate the effectiveness of our proposed CSI graph construction method,
we compare it against five intuitive methods: ``Amp", ``Phase", ``F-Amp", ``F-Phase", and ``F-A1P2". ``Amp" and ``Phase" utilize the preprocessed amplitude and phase directly to construct CSI graphs. ``F-Amp" and ``F-Phase" employ the fine-grained graph construction scheme based on amplitude and phase fusion methods, respectively.
In the amplitude fusion method, three channels of the CSI image are filled with the amplitude, the up-down flipped amplitude, and the left-right flipped amplitude.
The phase fusion method adopts a similar strategy.
``F-A1P2" adopts a scheme similar to our proposed fine-grained graph construction scheme. However, unlike the amplitude-phase fusion method detailed in Section~\ref{Amplitude-Phase Fusion}, it fills two RGB channels with amplitude and phase, while the third channel is filled with the up-down flipped phase.

We first evaluate these methods in Scenario 1, as shown in Fig.~\ref{csi_fusion_s10}. Methods utilizing amplitude information achieve both lower mean error and smaller error ranges compared to those using phase information. This is because the raw phase values actually collected by the wireless NIC are inaccurate~\cite{li2019af}, introducing additional errors. Moreover, the mean error and error range using the fine-grained graph construction methods are both lower than those using the original graph construction methods. This indicates that the proposed fine-grained graph construction method can enhance the CSI feature representation, thereby improving the localization performance. Among all methods, our proposed method achieves the best performance. 
Next, we focus on the fine-grained methods (i.e., ``F-Amp", ``F-Phase",  ``F-A1P2", and ``Proposed'') to specifically examine the effectiveness of the proposed amplitude-phase fusion method. In Scenario 1 with obstacle interference, the proposed method again achieves the best accuracy as shown in Fig.~\ref{csi_fusion_s11}.
	
We further compare the four methods under different bandwidths in Scenario 1, as shown in Fig.~\ref{CSI_fusion_BW}. Notably, the proposed method consistently achieves the lowest localization error across all bandwidths. This demonstrates that the amplitude-dominant, phase-assisted fusion strategy effectively combines the robustness of amplitude with the spatial resolution of phase, with especially pronounced improvements when the bandwidth becomes larger. Additionally, the localization error of ``F-Amp" at all three bandwidths is consistently higher than that of ``F-Phase", indicating that the amplitude is more stable. Although phase provides richer spatial information, it is easily affected by hardware noise and thus less robust when used alone. Finally, the superior performance of ``F-A1P2" over ``F-Phase" further highlights the benefit of incorporating amplitude to improve localization robustness.

\textbf{2) Effect of sampling densities.}
\begin{figure}[t]
  \centering
  \vspace{-2ex}
  \includegraphics[width=0.45\textwidth]{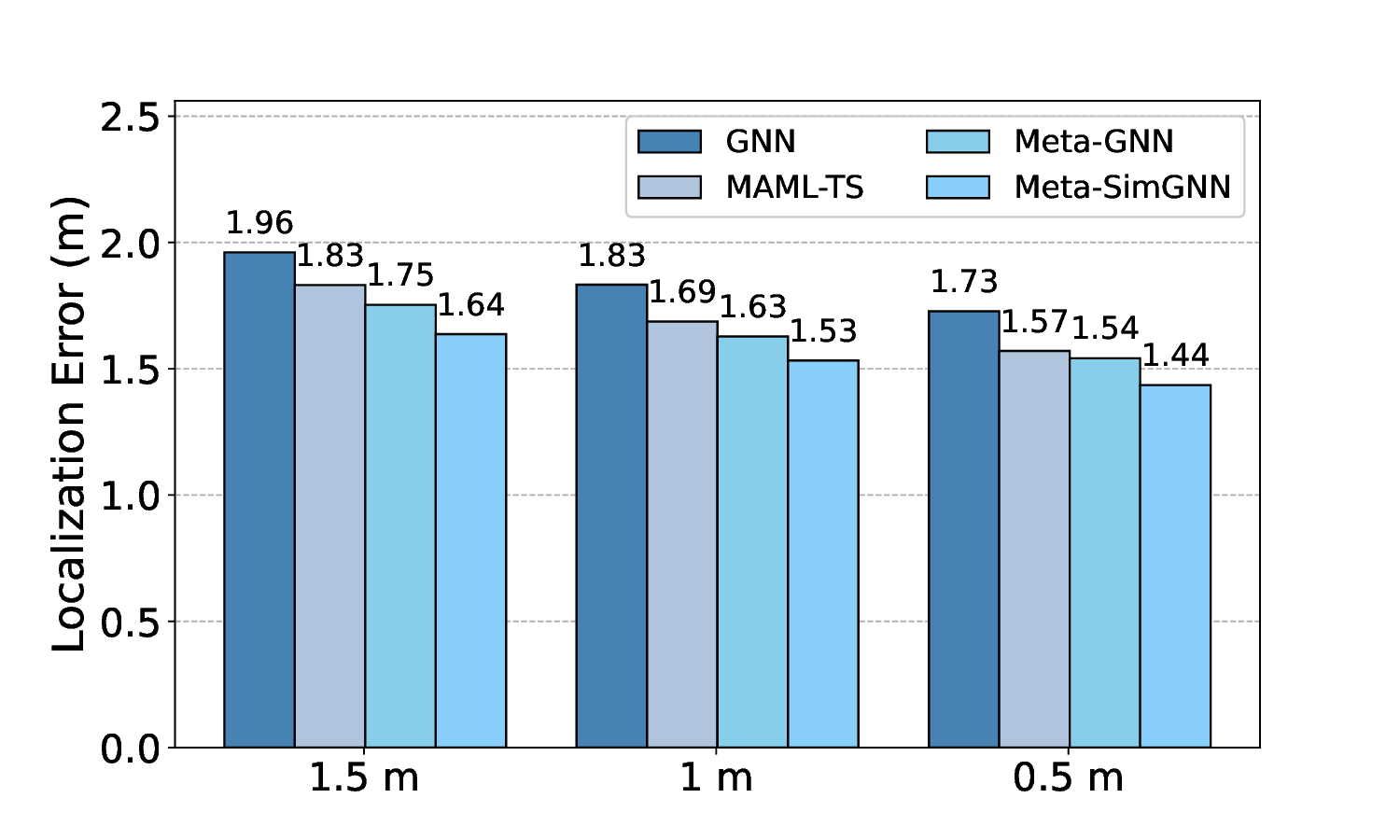}
  \vspace{-2ex}
  \caption{Localization performance under different sampling densities.}
  \label{sampleDensity}
  \vspace{-3ex}
\end{figure}
When collecting training samples, the UE collects CSI at each RP. The number of RPs deployed in a unit area is defined as the sampling density. Different sampling densities affect the number of training samples, which further affects the localization performance of the model. Therefore, it is essential to evaluate how sampling density impacts localization performance.
The sampling density is measured by the interval between RPs. Three RP intervals (i.e., 0.5~\!m, 1~\!m, and 1.5~\!m) are selected in Scenario 3.
Fig.~\ref{sampleDensity} displays the average localization error of GNN, MAML-TS, Meta-GNN, and \name at different RP intervals. It can be observed that the mean error gradually decreases as the RP interval decreases. That is to say, a smaller sampling interval provides more CSI samples and localization features, allowing the neural network model to see more training data, thus improving the estimation accuracy.
Notably, \name outperforms all other methods across each RP interval. Crucially, even under a large RP interval (e.g., 1~\!m), its average error is lower than that of the baseline method under a small RP interval (e.g., 0.5~\!m), demonstrating that the fine-grained graph construction and similarity-guided meta-learning strategy enable more effective exploitation of CSI features to improve localization performance. 
Moreover, MAML-TS, Meta-GNN, and \name all achieve lower mean errors than GNN, illustrating that the meta-learning framework can effectively adapt to achieve accurate localization using sparse CSI measurements.

\subsection{Generalization for Dynamic Device Configuration}

In the following, we aim to evaluate the performance of \name under dynamic device configurations, including different bandwidths, numbers of antennas, and numbers of APs.

\textbf{1) Performance under varying bandwidth.}
\begin{figure}[t]
  \centering
  \vspace{-2ex}
  \includegraphics[width=0.45\textwidth]{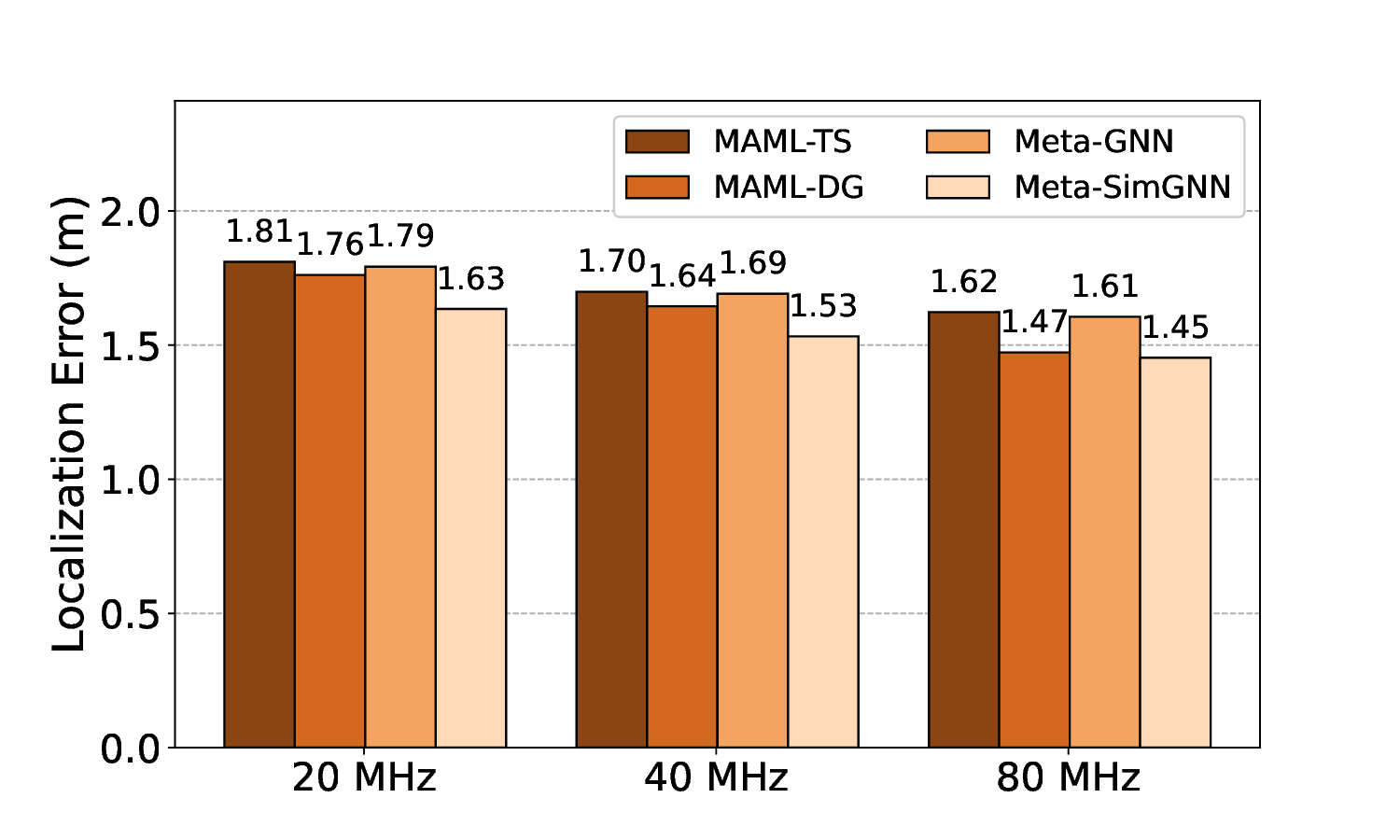}
  \vspace{-2ex}
  \caption{Localization error under varying bandwidths.}
  \label{badwidths}
  \vspace{-2ex}
\end{figure}
\begin{figure}[t]
	\centering
    \vspace{-2ex}
	\includegraphics[width=0.45\textwidth,trim=0 0 0 35pt,clip]{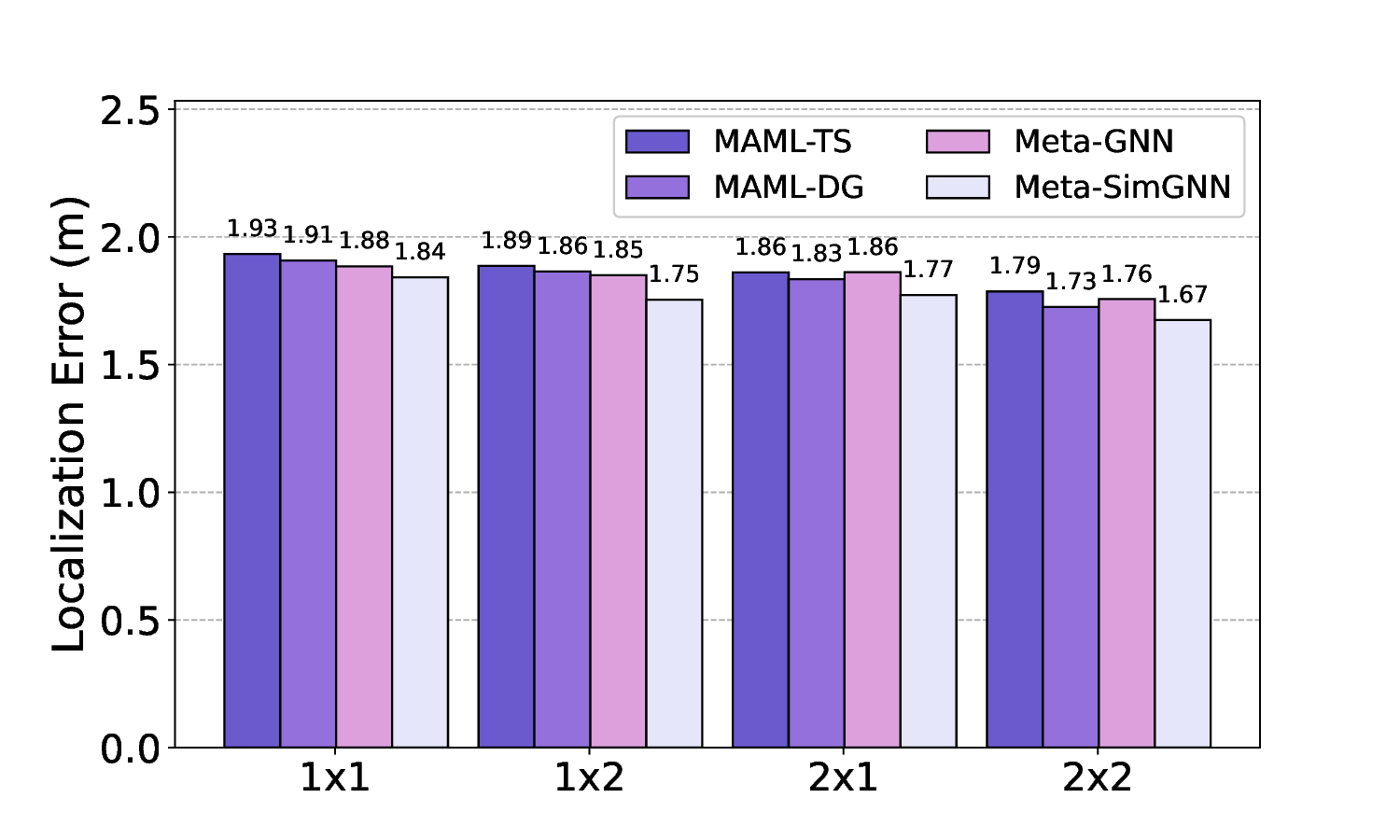}
    \vspace{-1ex}
	\caption{Localization performance under varying antenna configurations.}
	\label{bar_antennas}
	\vspace{-2ex}
\end{figure}
We first evaluate the localization performance under different bandwidths.
Here, we train separate models on Scenario 1, referring to Fig.~\ref{fig:scen1}, with bandwidths of 20, 40, and 80~\!MHz, and test each model only under its corresponding bandwidth in the scenario with obstacles interfering, referring to Fig.~\ref{fig:scen1_pass}.
As shown in Fig.~\ref{badwidths}, the mean error gradually decreases as the bandwidth increases across the four methods.
This is due to the fact that a larger bandwidth contains more subcarriers, providing richer CSI features to improve localization performance.
Among the four methods, the \name demonstrates the best localization performance across all bandwidth settings. This verifies that the \name can effectively utilize the available CSI under different bandwidths and obtain lower localization errors compared to the other three methods.

\textbf{2) Performance under varying antenna configurations.}
We then evaluate the localization performance under varying antenna configurations.
The model is trained using CSI collected in Scenario 1 with antenna configurations of 1$\times$1, 1$\times$2, 2$\times$1, and 2$\times$2, and tested using CSI collected in the presence of obstacles, with varying antenna configurations.
Fig.~\ref{bar_antennas} compares the average localization errors of MAML-TS, MAML-DG, Meta-GNN, and \name across four different antenna configurations. 
As antenna configuration scales from 1$\times$1 to 2$\times$2, the mean localization errors of all four methods decrease, owing to the enhanced spatial diversity in CSI measurements provided by additional transmit and receive antenna pairs.  
Notably, \name consistently outperforms the other three methods across different antenna configurations, because of its fine-grained CSI graph construction that effectively extracts and learns unique localization features, enabling robust adaptation to antenna configuration changes.

\begin{figure}[t]
  \centering
  \vspace{-1ex}
  \includegraphics[width=0.45\textwidth]{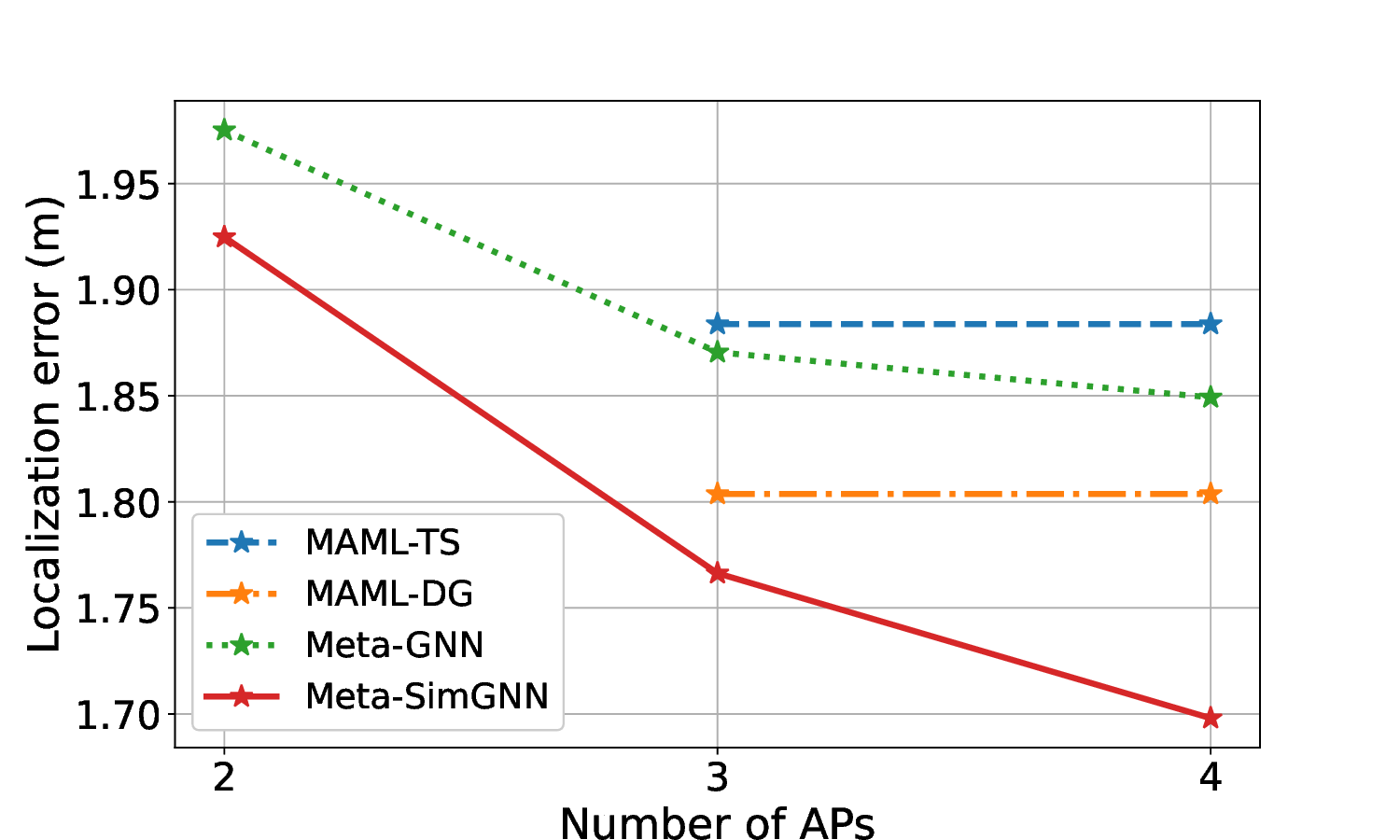}
  \vspace{-1ex}
  \caption{Localization errors under different numbers of APs.}
  \label{APnum_alg}
  \vspace{-1ex}
\end{figure}

\textbf{3) Performance under varying AP's number.} This experiment evaluates the effect of the number of APs on localization performance.
Fig.~\ref{APnum_alg} demonstrates the localization errors of the four localization methods with different numbers of APs.
For this experiment, CSI samples from Scenarios 1 and 3 are used for training, while samples from Scenario 2 are used for testing.
As depicted in the figure, the localization errors of the Meta-GNN and \name methods decrease as the number of WiFi APs increases.
This is because both methods leverage GNNs, which can fully utilize all available WiFi APs.
More available WiFi APs can provide more CSI and rich AP spatial topology information to improve localization performance.
Meanwhile, we can observe that the performance of MAML-TS and MAML-DG is lacking when the number of APs is less than 3, and it remains unchanged when it exceeds 3. This is because they are limited to utilizing three WiFi APs for localization. When the number of available WiFi APs is less than three, they do not work. When more than three APs are available, they cannot utilize the newly added CSI information, and the localization performance would not be improved with the AP's number.
In addition, the localization error of the proposed \name is the lowest among the four methods, verifying its superior localization performance.
\begin{figure}[t]
  \centering
  \includegraphics[width=0.45\textwidth,trim=0 0 0 35pt,clip]{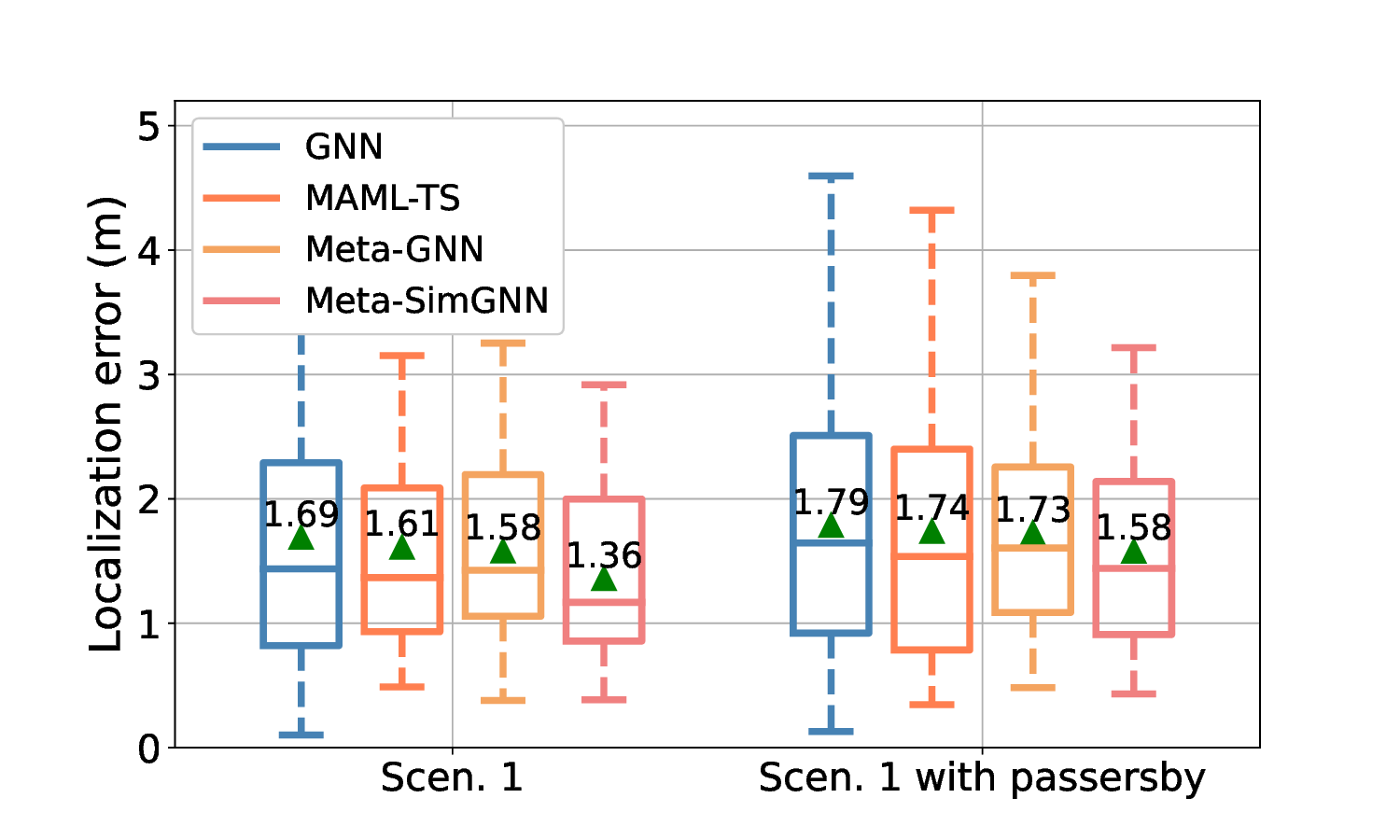}
  \vspace{-1ex}
  \caption{Localization performance in two environments.}
  \label{box_scenarios}
  \vspace{-1ex}
\end{figure}

\subsection{Generalization Across Scenarios}

In the following, we aim to evaluate the performance of \name across scenarios.

\textbf{1) Effect of newly added obstacles.}
Fig.~\ref{box_scenarios} compares the localization performance of the GNN, MAML-TS, Meta-GNN, and \name methods in the presence of newly added obstacles.
Here, we train the model using the CSI from Scenario 1 and test it in Scenario 1 with passersby.
As shown in the figure, the four methods achieve larger localization errors in the dynamic environment of Scenario 1 with passersby.
This can be attributed to the difference between the training and testing datasets.
When the test data is collected from the same environment (Scenario 1) without any interference, the environmental conditions closely resemble those of the training data, resulting in better localization performance.
However, in the presence of dynamic interference caused by passersby, the CSI changes significantly, making it less similar to the training data and leading to increased localization errors.
Notably, the \name achieves the lowest localization errors among all methods in both environments, demonstrating its ability to better adapt to environmental changes, thereby improving generalization performance. Furthermore, all meta-learning-based methods (i.e., Meta-GNN, MAML-TS, and \name) outperform GNN, confirming the effectiveness of meta-learning in improving generalization under dynamic environmental conditions.

\begin{figure}[t]
  \centering
  \vspace{-2ex}
  \includegraphics[width=0.45\textwidth]{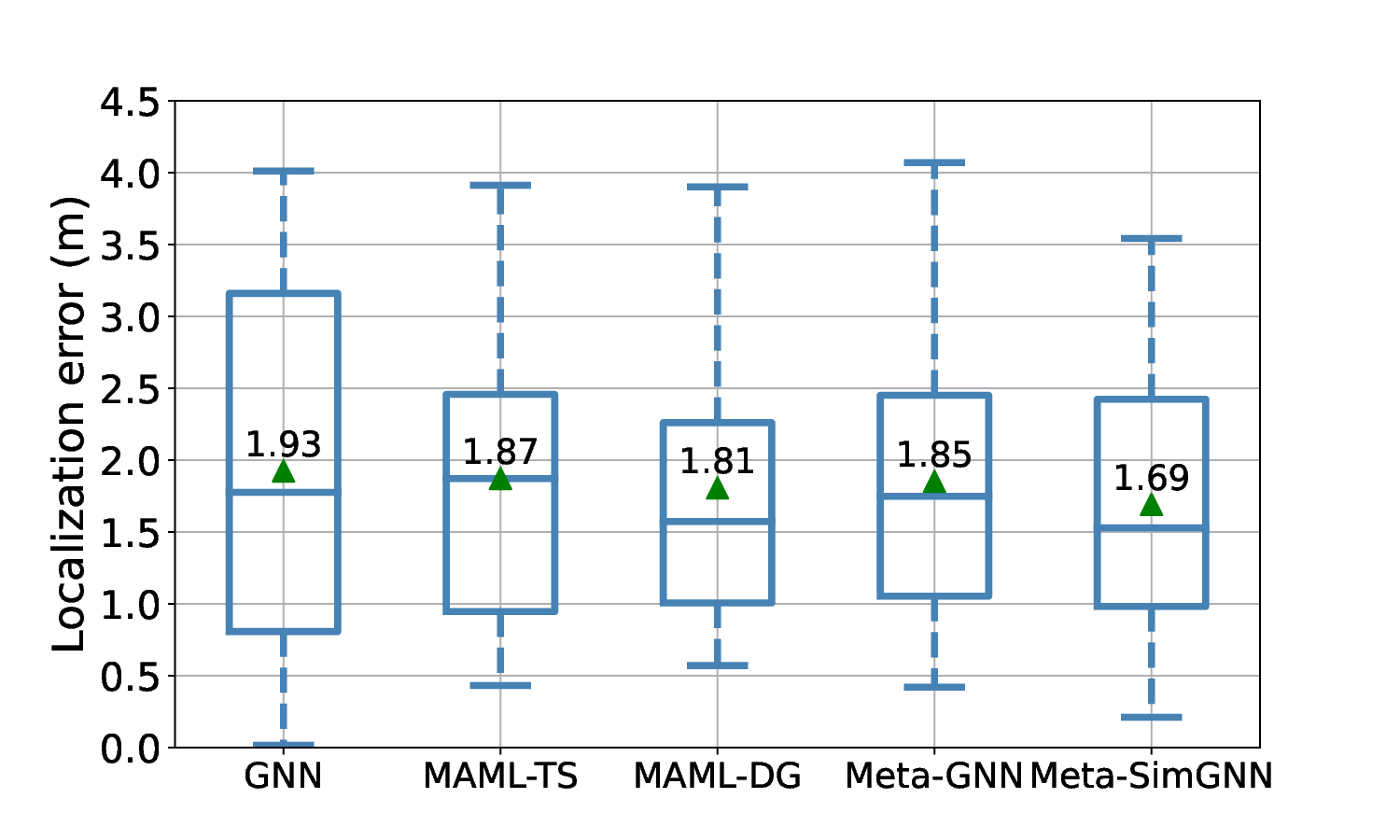}
  \vspace{-2ex}
  \caption{Localization performance of five methods across scenarios.}
  \label{box_algorithms}
  \vspace{-2ex}
\end{figure}

\begin{table}[t]
  \centering
  \caption{Localization Errors of Five Methods Across Scenarios}
  \large 
  \resizebox{\columnwidth}{!}{
  \begin{tabular}{|c|c|c|c|c|c|}
  \hline
  Methods & \name & Meta-GNN & MAML-DG & MAML-TS & GNN\\
  \hline
  Mean error (m) & 1.69 & 1.85 & 1.81 &1.87 &1.93 \\
  \hline
  STD (m) & 0.89 & 0.91 & 0.98 & 1.08 & 1.24 \\
  \hline
  \end{tabular}
  }
  \label{tab1}
  \vspace{-3ex}
\end{table}
\textbf{2) Performance across scenarios.}
We compare the generalization performance of the proposed \name and four baseline methods given in Section~\ref{Baseline}, i.e., GNN, Meta-GNN, MAML-TS, and MAML-DG.
Fig.~\ref{box_algorithms} shows the localization errors of five methods, adopting the CSI samples of Scenarios 1 and 3 for training and the samples of Scenario 2 for testing.
For a clearer comparison, the mean error and standard deviation (STD) of the five methods are summarized in Table~\ref{tab1}.

The results illustrate that the GNN has the largest mean error and error range, indicating its inability to effectively adapt to cross-scenario variations, such as changes in the layout of the localization space layout.
In contrast, the MAML-TS, MAML-DG, Meta-GNN, and \name methods reduce the mean error and the error range compared to the GNN, validating the role of meta-learning in improving cross-scenario generalization performance.
Moreover, Meta-GNN outperforms MAML-TS, showing that meta-learning combined with the GNN can fully utilize the AP spatial topology information to improve the localization performance. However, the performance of Meta-GNN is still not good enough.
We can see that \name has a smaller localization error than the other four methods, verifying its superior generalization performance across scenarios. This improvement can be attributed to the enhanced fine-tuning process in \name, which incorporates scenario similarity to improve cross-scenario localization performance.

\textbf{3) Effect of the dataset size for fine-tuning.}
\begin{figure}[t]
  \centering
  \includegraphics[width=0.45\textwidth,trim=0 0 0 35pt,clip]{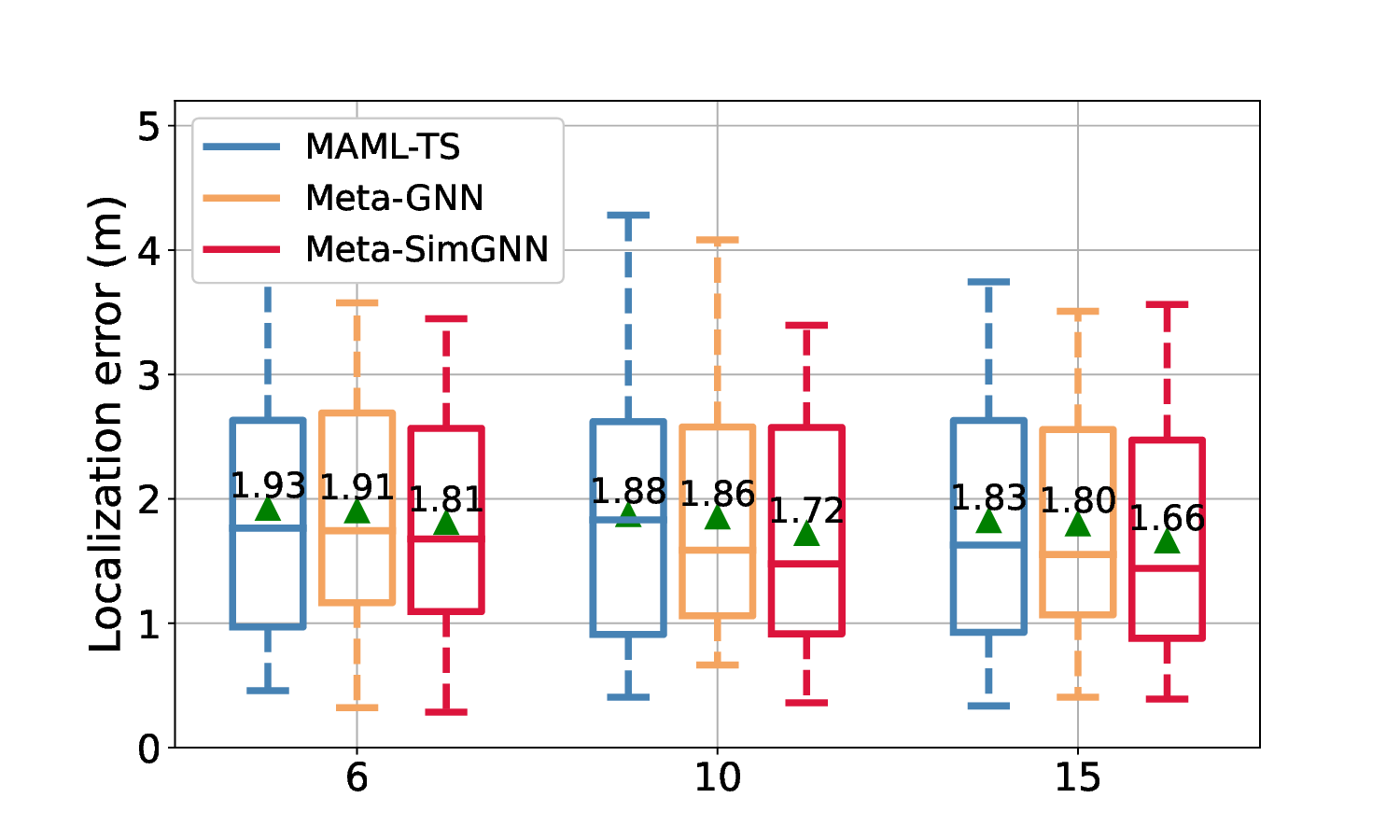}
  \vspace{-2ex}
  \caption{Localization performance of the number of fine-tuned samples at each RP.}
  \label{sanples_num}
  \vspace{-2ex}
\end{figure}
Since we perform fine-tuning with a limited number of samples, the number of samples impacts the localization performance. 
To investigate this, we train the model using CSI from Scenario 3 and fine-tune it with CSI from Scenario 2. Fig.~\ref{sanples_num} illustrates the localization performance of MAML-TS, Meta-GNN, and \name under different numbers of samples used for fine-tuning at each RP. 
The results demonstrate that increasing the number of samples for fine-tuning reduces both the mean localization error and the error range. This is because a larger fine-tuning dataset enables the model to learn the unique localization features of the new scenarios more comprehensively, thereby improving location estimation accuracy. 
It is noteworthy that \name still outperforms MAML-TS and Meta-GNN under conditions with only a few samples (e.g., 6 samples per RP) adopted for fine-tuning. This is due to the fine-grained CSI graph construction and similarity-guided fine-tuning strategy, which efficiently extract and learn the unique localization features from limited samples of the new scenario, enabling rapid and accurate localization.

\subsection{Real-World Experiment}
\begin{figure}[t]
  \centering
  \vspace{-2ex}
  \includegraphics[width=0.45\textwidth]{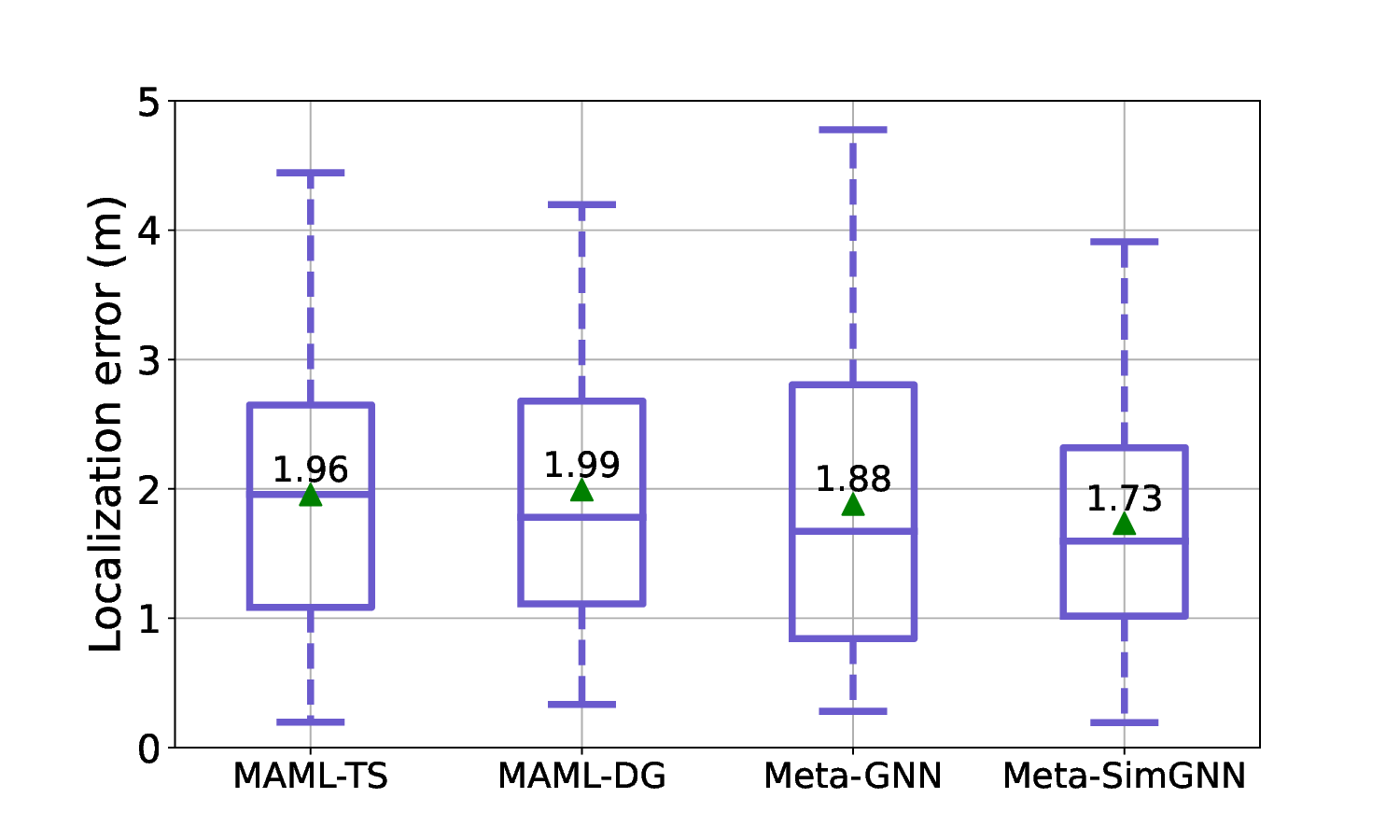}
  \vspace{-2ex}
  \caption{Localization performance in a real-world setting.}
  \label{box_mixMHz}
  \vspace{-2ex}
\end{figure}

Now, we evaluate the performance of \name in a real-world setting, where both the number of APs and the bandwidth vary dynamically.
Specifically, we use the CSI samples from Scenarios 1 and 3 for meta-training and the part CSI samples from Scenario 2 for fine-tuning. The fine-tuned model is then tested under a random number of APs with the bandwidth of each AP randomly selected from the set $\{20,40,80\}$~\!MHz.
Under this setup, Fig.~\ref{box_mixMHz} presents the localization performance of the four methods. It can be seen that the proposed \name shows the best performance among the four methods.
This is because \name can fully utilize the available WiFi APs and extract rich CSI features, enabling it to adapt well to environmental variations and AP heterogeneity.
Furthermore, the Meta-GNN method obtains a lower mean localization error compared to MAML-TS and MAML-DG. This is because Meta-GNN utilizes all available APs for localization, whereas MAML-TS and MAML-DG are limited to three APs. Moreover, MAML-TS and MAML-DG do not take into account the impact of variations in the bandwidth, resulting in lower localization performance.

\section{Conclusion}\label{Conclusion}
In this paper, we have proposed \name, an adaptive and robust WiFi indoor localization system designed to address the challenges of environmental variations and the dynamic configurations of WiFi APs.
We first have introduced GNNs to effectively leverage the spatial topological information of the APs to cope with the variations of APs' location and number.
Then, we have proposed a fine-grained CSI graph construction scheme to improve the localization robustness. It consists of two parts: an amplitude-phase fusion method and a feature extraction method.
The amplitude-phase fusion method exploits CSI in both the frequency and spatial domains, fusing amplitude and phase into CSI images to enhance data reliability.
Meanwhile, the feature extraction method utilizes spatial pyramid pooling to generate bandwidth-independent and dimension-consistent features, thereby resolving the dimensionality issue.
To improve adaptability across scenarios, we have designed a similarity-guided meta-learning strategy. It selects the scenario-specific meta-parameters as the initial model parameters for the fine-tuning stage by comparing the similarity between the new scenario and historical scenarios.
Experimental results have demonstrated that \name effectively tackles environmental variations and dynamic device configurations, significantly improving localization generalization and robustness.
In future work, we plan to evaluate our proposed \name in larger-scale, AP-dense real-world environments, and with heterogeneous devices, to further investigate its generalization and robustness.

\bibliographystyle{IEEEtran}
\bibliography{myrefe2}

\end{document}